\documentclass{article}

\PassOptionsToPackage{numbers, compress}{natbib}

 \usepackage[preprint]{neurips_2026}

\usepackage[utf8]{inputenc} 
\usepackage[T1]{fontenc}    
\usepackage{hyperref}       
\usepackage{url}            
\usepackage{booktabs}       
\usepackage{amsfonts}       
\usepackage{nicefrac}       
\usepackage{microtype}      
\usepackage{xcolor}         

\usepackage{amsmath}
\usepackage{wrapfig}
\usepackage{subfig}
\usepackage{xspace}
\newcommand{\ourmodel}{GARD\xspace}
\usepackage{graphicx}
\usepackage{multirow}
\usepackage{lipsum}

\title{Geometry-Aware Representation Denoising for \\Robust Multi-view 3D Reconstruction}

\author{
\textbf{Jin Hyeon Kim}$^{1\ast}$ 
\quad \textbf{Jaeeun Lee}$^{1\ast}$ 
\quad \textbf{Claire Kim}$^1$ 
\\
\textbf{Kyoungjin Oh}$^1$ 
\quad \textbf{Paul Hyunbin Cho}$^1$
\quad \textbf{Jaewon Min}$^1$ 
\quad \textbf{Yeji Choi}$^1$ 
\\
\textbf{Jihye Park}$^2$ 
\quad \textbf{Hyunhee Park}$^{2}$  
\quad \textbf{Minkyu Park}$^2$
\quad \textbf{Seungryong Kim}$^{1\dagger}$ 
\\[0.5em]
{
$^1$KAIST AI 
\qquad $^2$Samsung Electronics 
}
\\[0.5em]
{\tt \href{https://cvlab-kaist.github.io/GARD/}{\textcolor{purple}{https://github.com/cvlab-kaist/GARD}}}
}

\begin{document}

\begingroup
\renewcommand{\thefootnote}{}
\footnotetext{$^\ast$: Equal contribution}
\footnotetext{$^\dagger$: Corresponding authors}
\endgroup

\maketitle

\begin{figure}[h]
    \centering
    \vspace{-10pt}
    \includegraphics[width=\linewidth]{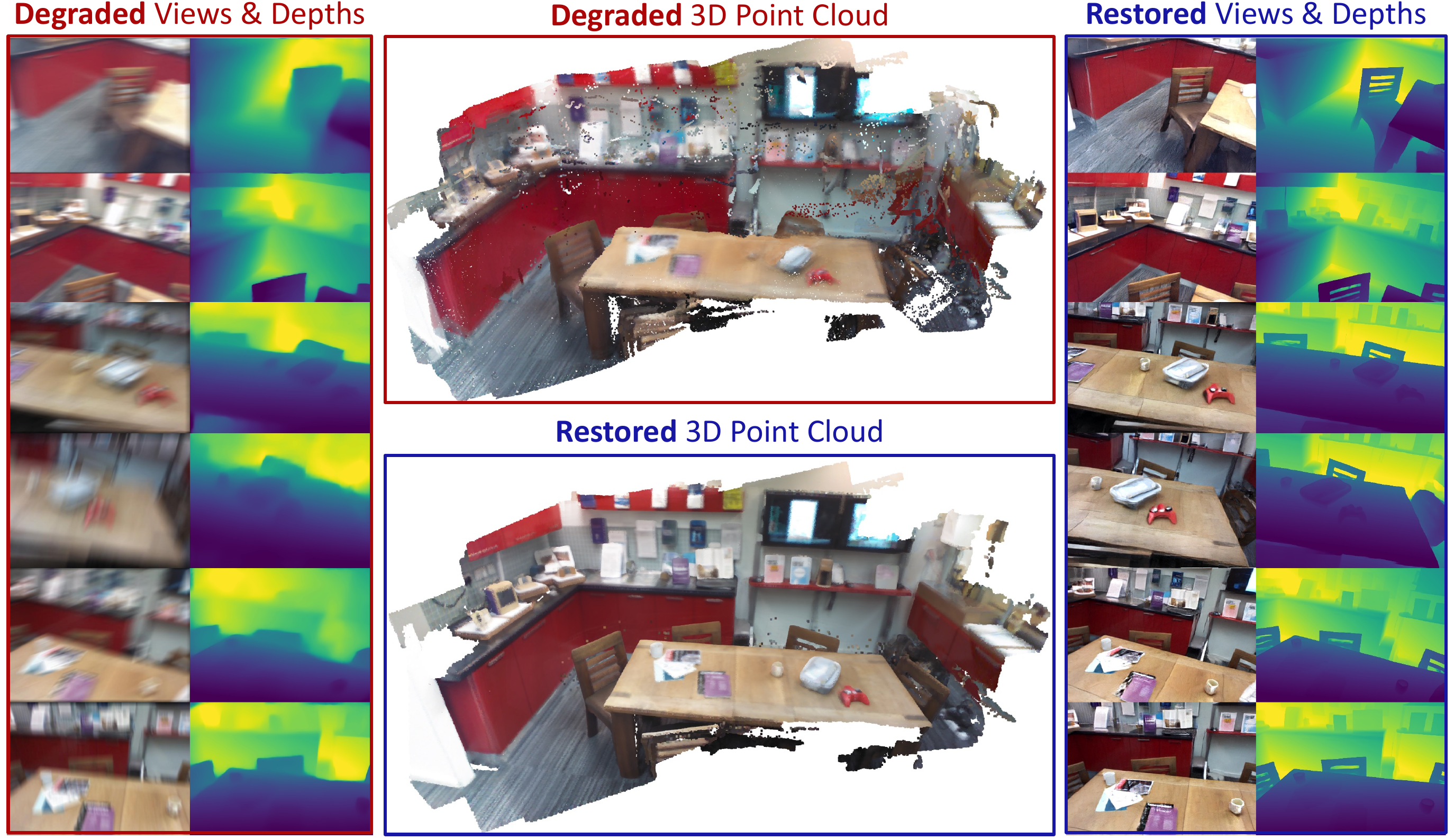}
    \vspace{-10pt}
    \caption{
    \textbf{Geometry-Aware Representation Denoising (GARD) framework.} Given degraded multi-view input images, our approach performs denoising on geometry-aware representations, thereby enabling simultaneous recovery of accurate 3D scene geometry and high-quality multi-view imagery.
    }
    \label{fig:teaser}
\end{figure}
\begin{abstract}

Multi-view 3D reconstruction has achieved remarkable progress with the advent of feed-forward 3D reconstruction models. However, these models are typically trained and evaluated under ideal, degradation-free imaging conditions, whereas real-world observations often contain degradations that differ significantly from such settings. Improving robustness for multi-view 3D reconstruction under degraded conditions therefore remains an important challenge. We present \textbf{Geometry-Aware Representation Denoising (GARD)}, a novel framework that performs diffusion-based multi-view restoration directly in the feature space of a feed-forward 3D reconstruction model. This design exploits the \textit{geometry-aware feature representations} of the 3D reconstructor to effectively recover accurate scene geometry. Furthermore, by employing an additional RGB image decoder, the refined representations can also be used to restore high-quality RGB images, thereby enabling the simultaneous recovery of 3D scene geometry and high-quality imagery. Comprehensive experiments on the Depth Anything 3 (DA3) benchmark demonstrate the effectiveness of the proposed GARD framework. 



\end{abstract}

\section{Introduction}
\label{sec:intro}

Reconstructing the 3D structure of a scene from 2D observations, commonly formulated as multi-view 3D reconstruction~\cite{hartley2003multiple, schonberger2016structure, furukawa2010towards, rosinol2020kimera, cramariuc2022maplab}, is a foundational problem in computer vision. It unlocks a wide range of real-world applications, including autonomous navigation~\cite{ceccarelli2022rgb, vargas2021overview, zhu2024llava}, robotics~\cite{kim2015real, lingbot-depth2026, 9506550}, and augmented and virtual reality~\cite{park2011handling, wloka1996interactive, okumura2006augmented}. Recent feed-forward reconstruction models~\cite{depthanything3, wang2025vggt, wang2025pi, wang2024dust3r, leroy2024grounding} have significantly advanced this task by replacing traditional multi-stage 3D reconstruction pipelines~\cite{hartley2003multiple, schonberger2016structure, 
furukawa2010towards} with end-to-end architectures that directly infer scene geometry from multi-view inputs. Built on transformer architectures~\cite{vaswani2017attention, dosovitskiy2020image}, their attention mechanism encode cross-view information to learn \textit{geometry-aware} representations, enabling accurate and scalable reconstruction under ideal imaging conditions.


However, real-world multi-view observations often deviate from this ideal setting. In practice, captured images and video sequences frequently suffer from degradations such as motion blur induced by camera motion~\cite{nah2017deep, rim2020real, malyugina2025unsupervised, liu2024boosting, han2025emergent}. These effects obscure fine textures and structural cues essential for reliable feature extraction and cross-view matching. As a result, the learned representations become less discriminative, disrupting geometric consistency across views. Since feed-forward models~\cite{depthanything3, wang2025vggt, wang2025pi, wang2024dust3r, leroy2024grounding} directly infer scene geometry from these features in a single forward pass, they lack mechanisms to explicitly correct such errors, allowing them to propagate through the network and accumulate in the final reconstruction. Improving robustness to such imperfect inputs therefore remains a key challenge for achieving reliable and consistent performance in multi-view 3D reconstruction.

\begin{figure}[t]
  \centering
  \includegraphics[width=\textwidth]{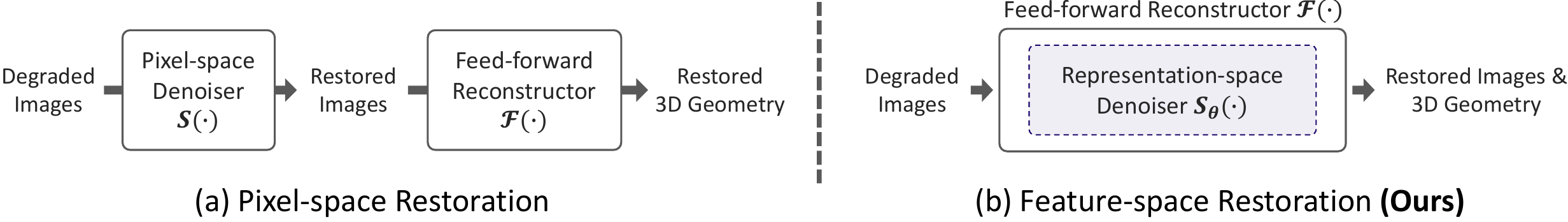}
\vspace{-10pt}
\caption{\textbf{Comparison of restoration denoising spaces.}
(a) A {restore-then-reconstruct} pipeline first performs pixel-space restoration prior to 3D reconstruction. However, performing restoration in a single-view setting~\cite{zamir2022restormer, conde2024high, chen2023hierarchical} or within a heavily compressed VAE-based latent space~\cite{mao2025sir, kingma2013auto} fails to preserve cross-view consistency and fine-grained geometric details, which often results in suboptimal geometric reconstruction. (b) In contrast, our \textbf{Geometry-Aware Representation Denoising (GARD)} operates on geometry-aware latent representations within a feed-forward reconstruction model, enabling the joint recovery of restored images and consistent 3D geometry across views.}
\label{fig:comparison}
\end{figure}

A key design question is where restoration should be performed in the 3D reconstruction pipeline. A straightforward approach would be to adopt a \textit{restore-then-reconstruct} paradigm (Fig.~\ref{fig:comparison} (a)), where degraded inputs are first restored in image space using existing image restoration models~\cite{zamir2022restormer, conde2024high, chen2023hierarchical, liang2024vrt, Youk_2024_CVPR, mao2025sir, zamfir2024complexityexperts} before being passed to the feed-forward reconstructor. However, current image restoration models are predominantly designed for single-view restoration and thus fail to leverage multi-view information and cannot enforce cross-view geometry consistency during restoration. While a recent multi-view restoration approach~\cite{mao2025sir} partially addresses this issue, it operates in a heavily compressed VAE-based latent space~\cite {kingma2013auto}, where information bottlenecks hinder the preservation of fine-grained details and geometric fidelity. Consequently, existing approaches remain suboptimal for multi-view image restoration and 3D reconstruction.

On the other hand, recent advances in Representation Autoencoders (RAEs)~\cite{zheng2025diffusion, tong2026scaling, kumar2026learning} further highlight the inherent shortcomings of conventional latent spaces. In particular, the compressed representations learned by VAEs~\cite{kingma2013auto, yao2025reconstruction} introduce information bottlenecks that hinder the preservation of fine-grained details and structural fidelity, which are essential for accurate multi-view 3D reconstruction. RAEs address this limitation by adopting high dimensional, semantically rich latent representations that better retain both global structure and local details, while maintaining a dedicated decoder for reconstructing high fidelity images. Motivated by this insight, we move beyond conventional image space~\cite{zamir2022restormer, chen2023hierarchical, conde2024high, liang2024vrt, Youk_2024_CVPR} and VAE-based formulations~\cite{mao2025sir} and instead exploit the geometry-aware feature space inherently encoded by feed-forward reconstruction models as a more suitable domain for denoising, while enabling image recovery through an auxiliary decoder.

To this end, we propose \textbf{Geometry-Aware Representation Denoising (GARD)}, a novel framework that learns a diffusion-based multi-view restoration denoiser model operating directly within the \textit{geometry-aware feature space} of a feed-forward reconstructor (Fig.~\ref{fig:comparison} (b)). By conducting denoising in this feature space, the proposed approach exploits high-dimensional representations inherently structured for scene geometry estimation, as well as the cross-view consistency encoded by feed-forward reconstruction models. This design preserves geometric fidelity while mitigating the information bottlenecks associated with VAE-based latent spaces and the inconsistencies introduced by image-space restoration. Furthermore, we adopt a dedicated image decoder~\cite{jang2026gld} to reconstruct high-quality RGB images from the refined representations, thereby enabling the joint recovery of high-quality imagery and accurate 3D scene geometry within a unified framework.

We validate our approach through extensive experiments on the Depth Anything 3 benchmark~\cite{depthanything3}, where controlled degradations are introduced to establish a rigorous evaluation protocol for restoration and reconstruction under motion blur degradation. To enable a fair comparison, we train the multi-view diffusion restoration model in both the geometry-aware feature space of the feed-forward reconstructor~\cite{depthanything3} and a conventional VAE-based latent space~\cite{kingma2013auto}, thereby isolating the impact of representation choice. We further compare our approach against restore-then-reconstruct pipelines for image restoration and scene geometry recovery, as well as dedicated image restoration methods evaluated using standard image quality metrics. Experimental results demonstrate that operating in the geometry-aware feature space yields improved geometric fidelity and visual quality, resulting in strong performance across pose estimation, 3D reconstruction, and image restoration benchmarks.

\section{Related Work}
\label{sec:rel_work}

\paragraph{Robust multi-view 3D reconstruction.}
The advent of feed-forward 3D reconstruction models~\cite{wang2024dust3r, leroy2024grounding, wang2025vggt, wang2025pi, depthanything3} has substantially advanced the field of multi-view 3D reconstruction. These models enable direct inference of scene geometry from multi-view images in a single forward pass, effectively replacing conventional multi-stage optimization pipelines~\cite{hartley2003multiple, schonberger2016structure, Schonberger_2016_CVPR, furukawa2010towards, furukawa2015mvs, DeTone_2018_CVPR_Workshops, an2024cross, 4587673, Sun_2021_CVPR}. Nevertheless, their performance degrades significantly in real-world settings~\cite{nah2017deep, rim2020real, malyugina2025unsupervised, liu2024boosting, han2025emergent}, where observations often contain noise such as distractors and degradations, since these models are primarily trained under ideal conditions with clean inputs. Prior works~\cite{han2025emergent, panvisual} have addressed robust multi-view 3D reconstruction in the presence of distractors. For example, RobustVGGT~\cite{han2025emergent} introduces an outlier rejection mechanism to eliminate irrelevant distractor views, while VGTW~\cite{panvisual} learns a dedicated distractor prediction head to identify and suppress distractor objects during reconstruction. In contrast, our approach is orthogonal to these methods, as we focus on degradations rather than distractors. In particular, camera motion blur is one of the most common and practically important degradations, arising frequently from handheld capture and dynamic imaging conditions. Such blur severely distorts fine textures, edges, and structural details, leading to unreliable geometric correspondence estimation and significantly hindering accurate 3D reconstruction.


\paragraph{Multi-view image restoration.}
Image restoration~\cite{Jiang_2025, chen2026lovif2026challengerealworld, zamir2022restormer, Youk_2024_CVPR, zuo2018convolutional, nah2018deepmultiscaleconvolutionalneural} aims to recover clean images from degraded observations, including deblurring~\cite{liang2024vrt, zamir2022restormer, chen2023hierarchical}, denoising~\cite{zamir2022restormer, zhang2018ffdnet, zamfir2024complexityexperts}, and super-resolution~\cite{lin2024diffbir, duan2025dit4sr, min2025text, kim2025unifieddiffusiontransformerhighfidelity}. Early CNN-based methods~\cite{zuo2018convolutional, zhang2018ffdnet} were later surpassed by transformer-based approaches~\cite{zamir2022restormer, chen2023hierarchical} such as Restormer~\cite{zamir2022restormer} and Hi-Diff~\cite{chen2023hierarchical}, which better capture long-range dependencies. InstructIR~\cite{conde2024high} further enables unified restoration through language conditioning. However, these methods operate on single images and cannot leverage multi-view complementary information for effective restoration. Although video restoration models such as VRT~\cite{liang2024vrt} and FMA-Net~\cite{Youk_2024_CVPR} process multi-frame inputs, they are predominantly trained on temporally adjacent video sequences and rely heavily on temporal coherence, limiting their applicability to multi-view scenarios with substantial viewpoint variation. Furthermore, while SIR-Diff~\cite{mao2025sir} introduces a sparse multi-view diffusion-based restoration framework, it operates in compressed VAE-based latent spaces~\cite{kingma2013auto} which may discard fine-grained visual structures. These limitations motivate restoration within geometry-aware representation spaces that preserve both cross-view consistency and detailed scene structure.




\paragraph{Representation space learning.}
Large-scale pretraining has made semantically rich visual representations a core component of modern vision systems~\cite{oquab2023dinov2, he2022masked, radford2021learning, tschannen2025siglip, lee2023domain}. Pretrained encoders produce high-dimensional feature spaces that generalize effectively across diverse downstream tasks~\cite{zheng2025diffusion, jang2026gld, tong2026scaling, kumar2026learning}. In generative modeling, latent diffusion models (LDMs)~\cite{rombach2022high, esser2024scaling, peebles2023scalable, kilian2024computational, kim2025unifieddiffusiontransformerhighfidelity, min2025text} improve efficiency by operating in compact VAE-based latent spaces~\cite{kingma2013auto}. However, since these latents are optimized mainly for reconstruction, they often lack rich semantic structure and geometric consistency, creating information bottlenecks for downstream reasoning. Representation Autoencoders (RAEs)~\cite{zheng2025diffusion, tong2026scaling, kumar2026learning} address this by replacing the VAE encoder with a frozen pretrained representation network, yielding richer latent spaces for diffusion. Similarly, feed-forward 3D reconstruction models~\cite{depthanything3, wang2025vggt, leroy2024grounding, wang2024dust3r, wang2025pi} learn strong geometric multi-view feature representations through transformer-based attention mechanisms, enabling representation spaces optimized for cross-view reasoning and scene geometry inference. Motivated by these advances, our approach performs diffusion-based feature restoration directly in such geometry-aware representation spaces, leveraging the already optimized geometric representations to better preserve and restore geometric feature representations during denoising while avoiding the limitations of VAE-based formulations.

\section{Method}
We propose a novel framework that performs denoising directly in the geometry-aware feature space of a frozen feed-forward 3D reconstruction model~\cite{depthanything3}. Our framework learns a multi-view latent diffusion model~\cite{wang2025ddt, zheng2025diffusion} to restore the degraded intermediate feature representations produced by the feed-forward reconstructor. This design enables simultaneous recovery of clean RGB images and accurate 3D scene geometry in a single forward pass with dedicated decoders~\cite{depthanything3, jang2026gld}, without retraining the underlying backbone.


\subsection{Task Formulation and Motivation}

Given a set of $V$ degraded multi-view images $\mathbf{I}_{\text{deg}} \in \mathbb{R}^{V \times H \times W \times 3}$, where $H$ and $W$ denote the image height and width respectively, \ our objective is to recover both the restored images $\mathbf{I}_{\text{res}} \in \mathbb{R}^{V \times H \times W \times 3}$ and the underlying 3D scene geometry $\mathbf{G} = \{\mathbf{G}_\text{depth}, \mathbf{G}_\text{pose}\}$. Here, $\mathbf{G}_\text{depth} \in \mathbb{R}^{V \times H \times W \times 1}$ denotes the per-view depth maps, and $\mathbf{G}_\text{pose} \in \mathbb{R}^{V \times 9}$ represents the corresponding camera pose parameters, consisting of translation and rotation quaternions.

\paragraph{Limitations of pixel-space denoising.}
As illustrated in Fig.~\ref{fig:comparison} (a), a straightforward solution is to adopt a {restore-then-reconstruct} pipeline, in which degraded inputs are first processed by a restoration denoiser $\mathcal{S}(\cdot)$ to obtain restored images, $\mathbf{I}_{\text{res}} = \mathcal{S}(\mathbf{I}_{\text{deg}})$, which are subsequently fed into the feed-forward reconstructor $\mathcal{F}(\cdot)$ for geometry estimation, i.e., $\mathbf{G} = \mathcal{F}(\mathbf{I}_{\text{res}})$. In practice, $\mathcal{S}(\cdot)$ can be instantiated using existing image restoration models~\cite{zamir2022restormer, chen2023hierarchical, conde2024high, zamfir2024complexityexperts, liang2024vrt, Youk_2024_CVPR}. Despite its conceptual simplicity, this pipeline inherits two fundamental limitations. First, the majority of restoration methods operate on single-view inputs, consequently, applying $\mathcal{S}(\cdot)$ independently to each view fails to leverage multi-view complementary information for effective restoration and cannot enforce cross-view geometric consistency due to its fundamental architecture design. This often leads to view-dependent artifacts and inconsistencies that propagate to the feed-forward 3D reconstructor, resulting in poor geometry estimation performance. 

Second, existing multi-view restoration methods remain constrained by both their modeling assumptions and underlying representation spaces. While video restoration models~\cite{liang2024vrt, Youk_2024_CVPR} can exploit complementary information across multiple views, these models are predominantly trained on temporally adjacent video frames and rely heavily on short-range temporal coherence, limiting their generalization to sparse multi-view scenarios. Furthermore, although a recent multi-view restoration approach~\cite{mao2025sir} explicitly models sparse cross-view interactions, it performs restoration in a heavily compressed VAE latent space~\cite{kingma2013auto}, thereby introducing an information bottleneck that removes fine-grained spatial structures and high-frequency details. Such information is critical for establishing accurate cross-view correspondences, and its absence can substantially degrade downstream 3D reconstruction performance. Collectively, these limitations hinder the effective exploitation of complementary multi-view information and restrict the preservation of geometric consistency and fine-grained scene structure during restoration.


\paragraph{Representation space perspective.}
Recent advances in Representation Autoencoders (RAEs)~\cite{zheng2025diffusion, tong2026scaling, kumar2026learning} further highlight the limitations of conventional compressed latent spaces such as VAEs, demonstrating that richer, high-dimensional representations are crucial for preserving structural and semantic information. This observation suggests that restoration performance is fundamentally tied to the expressiveness of the underlying representation. In our case, the intermediate feature space $\mathbf{z}$ of the feed-forward 3D reconstruction model $\mathcal{F}(\cdot)$ is inherently geometry-aware due to cross-view interactions encoded by the multi-view transformer encoder $\mathcal{E}(\cdot)$, making it a more suitable domain for multi-view denoising. These observations further motivate restoration mechanisms that operate directly in geometry-aware representation spaces, enabling the joint preservation of visual fidelity and multi-view consistency.

\begin{figure}[t]
  \centering
  \includegraphics[width=\textwidth]{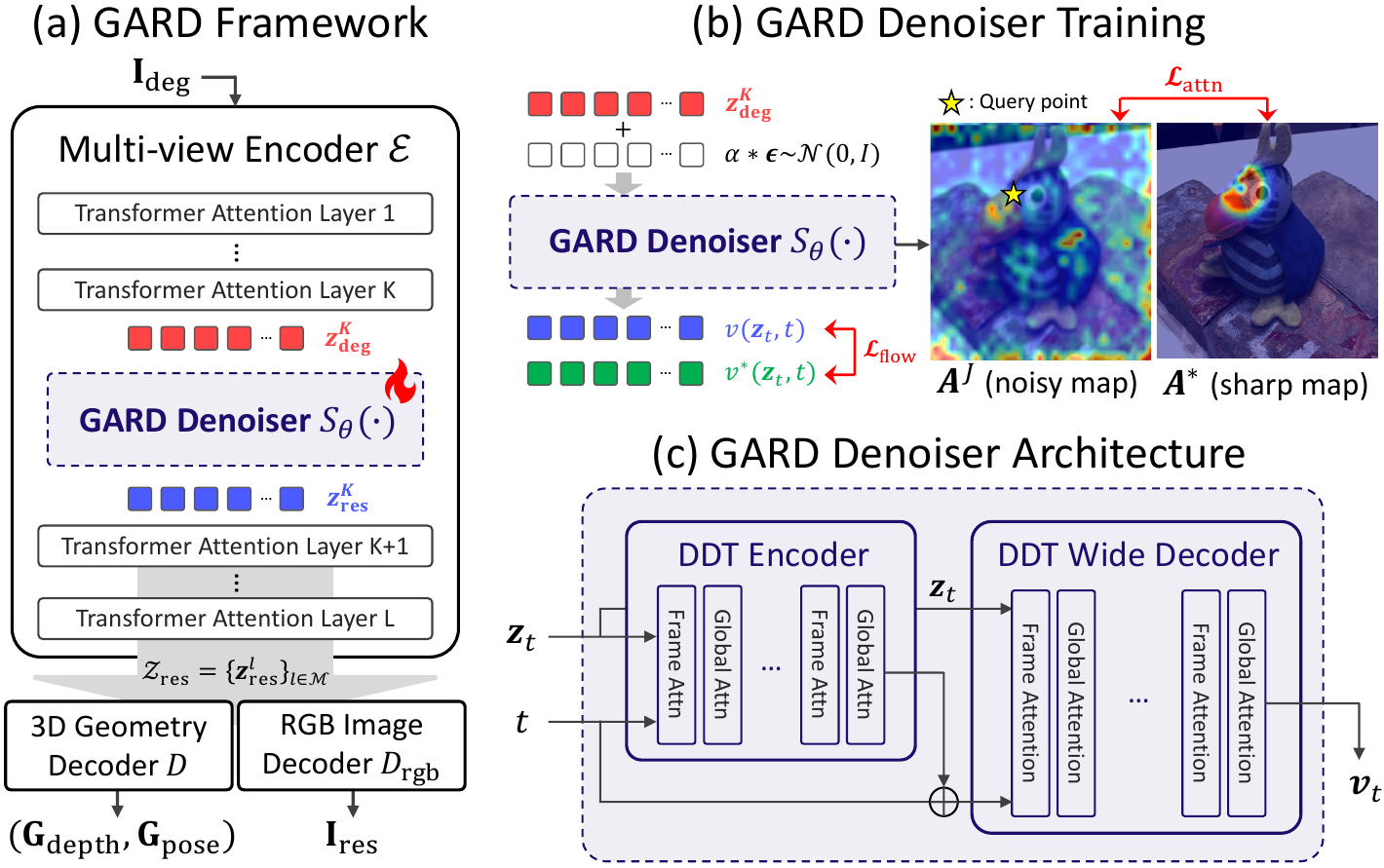}
  \vspace{-10pt}
\caption{\textbf{Overview of the GARD framework.}
(a) The GARD denoiser $\mathcal{S}_\theta(\cdot)$ is learned within the representation space of a frozen multi-view encoder~\cite{depthanything3} to restore degraded intermediate representations $\mathbf{z}_{\text{deg}}^K$ into restored representations $\mathbf{z}_{\text{res}}^K$ before they are propagated through the remaining encoder layers. The restored representations $\mathcal{Z}_{\text{res}}$ are then decoded by their respective decoders to produce geometry predictions and restored RGB images. (b) The GARD denoiser is optimized using an interpolated flow matching loss together with an attention alignment loss, which jointly learns the mapping from degraded to clean feature representations while preserving geometric consistency through explicit alignment of attention maps. (c) The GARD denoiser adopts a multi-view latent diffusion architecture~\cite{zheng2025diffusion}, comprising a DDT encoder~\cite{wang2025ddt} and a DDT wide decoder, with global attention layers inserted to enable multi-view modeling, thereby facilitating global context aggregation and reconstruction of high-dimensional multi-view representations.}
\label{fig:architecture}
\end{figure}
\begin{figure}[t]
\centering
\includegraphics[width=\textwidth]{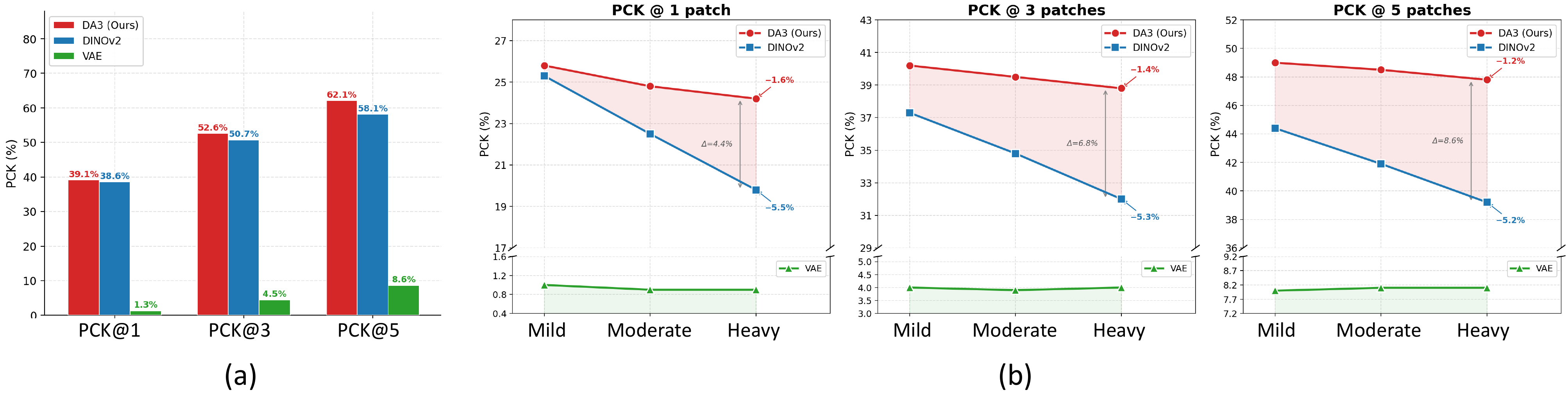}
\vspace{-15pt}
\caption{\textbf{Geometry-aware feature analysis conducted on ETH3D~\cite{schops2017multi}.} We evaluate the PCK accuracy of three feature cost volumes~\cite{kingma2013auto, oquab2023dinov2, depthanything3} under two experimental settings to validate the effectiveness of our proposed denoising space. (a) PCK performance on high-quality (HQ) input images. (b) PCK performance under progressively increasing levels of degradation (mild, moderate, and heavy), demonstrating robustness to input corruption. The correspondence visualizations for each cost volume are provided in Fig.~\ref{fig:suppl_cost_volume} of the supplementary material.}
\label{fig:geo_aware}
\end{figure}
\subsection{GARD: Geometry-Aware Representation Denoising}
\paragraph{Overview.}
To address the limitations of pixel-based and VAE-based latent-space restoration~\cite{mao2025sir, kingma2013auto}, we propose \textbf{Geometry-Aware Representation Denoising (GARD)} to perform denoising directly in the geometry-aware feature space of a feed-forward reconstructor $\mathcal{F}(\cdot)$. 
As illustrated in Fig.~\ref{fig:architecture} (a), the restoration denoiser $\mathcal{S}_{\theta}(\cdot)$, which we refer to as the GARD denoiser, operates directly within the feature representation space of a pretrained feed-forward reconstruction model $\mathcal{F}(\cdot)$. 

Specifically, the reconstructor $\mathcal{F}(\cdot)$ comprises a multi-view encoder $\mathcal{E}(\cdot)$ and a geometry decoder $\mathcal{D}(\cdot)$. The encoder $\mathcal{E}(\cdot)$ maps the input images to a latent feature representation $\mathbf{z} = \mathcal{E}(\mathbf{I})$, where $\mathbf{z} \in \mathbb{R}^{V \times N \times C}$, with $N$ denoting the number of tokens and $C$ the feature dimensionality. The geometry decoder $\mathcal{D}(\cdot)$ subsequently predicts the scene geometry from the latent representation, i.e., $\mathbf{G} = \mathcal{D}(\mathbf{z})$. Specifically, given degraded inputs $\mathbf{I}_{\text{deg}}$, they are first encoded by an $L$-layer multi-view encoder $\mathcal{E}(\cdot)$ to produce layer-wise latent representations $\{\mathbf{z}_{\text{deg}}^{l}\}_{l=1}^{L} = \mathcal{E}(\mathbf{I}_{\text{deg}})$. The GARD denoiser $\mathcal{S}_{\theta}(\cdot)$, implemented as a multi-view latent diffusion model, composed of transformer attention layers is inserted at the $K$-th layer of $\mathcal{E}$ to refine the degraded feature representation at layer $K$, yielding $\mathbf{z}_{\text{res}}^{K} = \mathcal{S}_{\theta}(\mathbf{z}_{\text{deg}}^{K})$. This refined representation is subsequently propagated through the remaining encoder layers, producing restored features $\{\mathbf{z}_{\text{res}}^{l}\}_{l=K}^{L}$. From these restored features, four feature levels $\mathcal{Z}_{\text{res}} = \{\mathbf{z}_{\text{res}}^{l}\}_{l \in \mathcal{M}}$, with $|\mathcal{M}| = 4$, are selected and provided as input to two task-specific decoders: a geometry decoder $\mathcal{D}(\cdot)$ for 3D geometry estimation and an RGB decoder $\mathcal{D}_{\text{rgb}}(\cdot)$ for image restoration. Note that the geometry decoder $\mathcal{D}(\cdot)$ is part of the feed-forward reconstructor, while the RGB image decoder $\mathcal{D}_{\text{rgb}}(\cdot)$ is adapted from \cite{jang2026gld} and fine-tuned for our framework. The final outputs are given by $\mathbf{G} = \mathcal{D}(\mathcal{Z}_{\text{res}})$ and $\mathbf{I}_{\text{res}} = \mathcal{D}_{\text{rgb}}(\mathcal{Z}_{\text{res}})$. By learning the GARD denoiser within the multi-view encoder and performing denoising directly in the geometry-aware feature space, the proposed framework enables simultaneous image restoration and 3D scene reconstruction in a single forward pass, without requiring separate restoration and reconstruction stages or retraining the underlying backbone.

\paragraph{Geometry-aware feature analysis.}
We first investigate the geometric encoding capability of feature representations produced by the multi-view encoder $\mathcal{E}(\cdot)$ to evaluate their effectiveness for representation-level denoising. Specifically, we adopt Depth Anything 3 (DA3) as our representative multi-view encoder and compare its feature representations with those of VAE~\cite{kingma2013auto} and DINOv2~\cite{oquab2023dinov2}. To this end, we measure keypoint correspondence accuracy using PCK by constructing feature-based cost volumes from the three different representations. Through two experiments, we demonstrate that DA3 encodes more geometry-aware representations than the alternatives. (1) Under clean high-quality (HQ) multi-view inputs, DA3 achieves the highest PCK across all three evaluation thresholds. (2) Under progressively increasing degradation levels (mild, moderate, and heavy), DA3 features exhibit superior robustness to input corruption. As shown in Fig.~\ref{fig:geo_aware}, the feature representations of DA3 consistently produce higher PCK scores and maintain stronger robustness across degradation levels compared to VAE- and DINOv2-based features. These results indicate that the multi-view encoder more effectively preserves geometric structure, making it more suitable for representation-level denoising and downstream 3D reconstruction tasks. Refer to Fig.~\ref{fig:suppl_cost_volume} in the supplementary material for qualitative visualizations of feature cost-volume correspondences for each representation.

\paragraph{Representation denoiser model.}
The GARD denoiser $\mathcal{S}_{\theta}(\cdot)$ is a multi-view diffusion architecture built on the $\text{DiT}^{\text{DH}}$ design from RAE~\cite{zheng2025diffusion}, which is well-suited for denoising in high-dimensional feature spaces. As illustrated in Fig.~\ref{fig:architecture} (c), it adopts an encoder and a wide decoder structure augmented with interleaved global attention layers to enable multi-view representation learning. Specifically, the model interleaves frame-level attention and global cross-view attention to aggregate contextual information both within and across views. Frame-level attention captures local spatial structure within each view, while global attention enables the model to exploit cross-view correspondences and enforce geometric consistency across multi-view feature representations.

\paragraph{Interpolated flow matching loss.}
The GARD denoiser $\mathcal{S}_{\theta}(\cdot)$ is trained to perform restoration on the $K$-th layer feature representation $\mathbf{z}_{\text{deg}}^{K}$ within the multi-view encoder $\mathcal{E}$. Unlike prior works that define flow trajectories from Gaussian noise to the clean representations~\cite{zheng2025diffusion, duan2025dit4sr}, we use the degraded latent itself as the source distribution, since $\mathbf{z}_{\text{deg}}^{K}$ already retains meaningful structural and geometric information beneficial for restoration. To improve robustness, we introduce a noise-perturbed source representation $\tilde{\mathbf{z}}_{\text{deg}}^{K} = \mathbf{z}_{\text{deg}}^{K} + \alpha \boldsymbol{\epsilon}$, where $\boldsymbol{\epsilon} \sim \mathcal{N}(0,\mathbf{I})$ and $\alpha \in [0,1]$ controls the perturbation magnitude. The resulting optimization objective follows the standard flow matching formulation:

\begin{equation}
\mathcal{L}_{\text{flow}}
=
\mathbb{E}_{t,
\mathbf{z}_{\text{deg}}^{K},
\mathbf{z}_{\text{clean}}^{K}}
\left[
\left\|
{v}(\mathbf{z}_t,t)
-
{v}^{*}(\mathbf{z}_t,t)
\right\|_2^2
\right].
\end{equation}

where $\mathbf{z}_t = (1-t)\tilde{\mathbf{z}}_{\text{deg}}^{K} + t\mathbf{z}_{\text{clean}}^{K}$, $t \sim \mathcal{U}(0,1)$ with the predicted velocity field as ${v}(\mathbf{z}_t,t) = \mathbf{v}_t = \mathcal{S}_{\theta}(\mathbf{z}_t,t)$ and ground-truth velocity field as ${v}^{*}(\mathbf{z}_t,t) = \mathbf{v}^*_t=\mathbf{z}_{\text{clean}}^{K} - \tilde{\mathbf{z}}_{\text{deg}}^{K}$. The restored representation is obtained by integrating the learned velocity field with an ODE solver, after which it is propagated through the remaining encoder layers and decoded for both geometry estimation and image restoration. An illustration is provided in Fig.~\ref{fig:architecture} (b).

\paragraph{Attention alignment loss.}
While the flow-matching objective enforces feature-level alignment, it does not explicitly encourage learning of cross-view correspondences. We therefore regularize the attention maps within the GARD denoiser $\mathcal{S}_{\theta}(\cdot)$ to align with geometrically consistent correspondence maps following prior attention alignment works~\cite{kwon2025cameo, nam2025emergenttemporalcorrespondencesvideo, jin2025matrixmasktrackalignment}. Specifically, we encourage the global attention weights for the $J$-th layer of the GARD denoiser to focus on geometrically corresponding regions rather than spurious artifacts. Let $\mathbf{A}^J \in \mathbb{R}^{V \times N \times N}$ denote the global attention map of the $J$-th layer within $\mathcal{S}_{\theta}(\cdot)$, and $\mathbf{A}^* \in \mathbb{R}^{V \times N \times N}$ denote the target correspondence maps obtained from the point cloud of clean multi-view inputs. The alignment loss is defined using cross-entropy as follows:
\begin{equation}
\mathcal{L}_{\text{attn}} = - \mathbb{E} \left[ \mathbf{A}^{*} \log \mathbf{A}^{J} \right].
\end{equation}
and is optimized jointly with the flow-matching objective, yielding the total loss $\mathcal{L} = \mathcal{L}_{\text{flow}} + \lambda_{\text{attn}} \mathcal{L}_{\text{attn}}$, where $\lambda_\text{attn}$ denotes the attention alignment loss coefficient. This supervision promotes sharper and more coherent attention patterns, thereby improving both reconstruction accuracy and structural fidelity. Further details and visualizations of the attention target correspondence maps and the effects of attention alignment are provided in Fig.~\ref{fig:suppl_attn_target} and Fig.~\ref{fig:suppl_attn} of the supplementary material.

\section{Experiments}
\label{sec:exp}

\subsection{Experimental Settings}
\label{sec:exp_setup}

\paragraph{Baselines.}
We compare our method against restore-then-reconstruct pipeline approaches, in which image restoration is performed prior to feed-forward 3D reconstruction, as illustrated in Fig.~\ref{fig:comparison}. The restoration stage is further divided into two categories: single-view restoration methods and multi-view restoration methods. For single-view restoration, we instantiate the restoration model using representative single-view image restoration models, including Restormer~\cite{zamir2022restormer}, HI-Diff~\cite{chen2023hierarchical}, InstructIR~\cite{conde2024high}, and MoCE-IR~\cite{zamfir2024complexityexperts}. These methods collectively cover conventional restoration architectures, diffusion-based approaches, and all-in-one image restoration frameworks. For multi-view restoration, we construct a baseline using the same denoising architecture as the proposed GARD denoiser, but operating directly in the VAE-based latent space~\cite{kingma2013auto} rather than in the geometry-aware feature space. We denote this baseline as VAE$_\text{MVD}$, where MVD refers to the Multi-View Denoiser. In addition, we include video restoration models, such as VRT~\cite{liang2024vrt} and FMA-Net~\cite{Youk_2024_CVPR}, as additional baselines due to their capability to process multi-view inputs. Results and discussions for another representative multi-view restoration model, SIR-Diff~\cite{mao2025sir}, are provided in the supplementary material.


\begin{table*}[t]
\caption{\textbf{Quantitative pose estimation results.} We report AUC5$\uparrow$ and AUC30$\uparrow$ for camera pose estimation on the DA3 benchmark~\cite{depthanything3}, comparing single-view and multi-view restoration approaches. The best result is highlighted in \textbf{bold} and the second best is \underline{underlined}.}
\label{tab:pose}
\centering

\resizebox{\linewidth}{!}{
\begin{tabular}{lcccccccccccccccccccc}
\toprule
\multirow{2}{*}{\textbf{Model}}
& \multicolumn{2}{c}{\textbf{HiRoom}}
& \multicolumn{2}{c}{\textbf{ETH3D}}
& \multicolumn{2}{c}{\textbf{DTU}}
& \multicolumn{2}{c}{\textbf{7Scenes}}
& \multicolumn{2}{c}{\textbf{ScanNet$++$}} \\
\cmidrule(lr){2-3} \cmidrule(lr){4-5} \cmidrule(lr){6-7} \cmidrule(lr){8-9} \cmidrule(lr){10-11}

& AUC5$\uparrow$ & AUC30$\uparrow$
& AUC5$\uparrow$ & AUC30$\uparrow$
& AUC5$\uparrow$ & AUC30$\uparrow$
& AUC5$\uparrow$ & AUC30$\uparrow$
& AUC5$\uparrow$ & AUC30$\uparrow$ \\
\midrule

\textit{Input Conditions} \\
\midrule
HQ Input
& 87.20 & 96.65
& 53.45 & 84.68
& 92.44 & 98.70
& 42.47 & 86.91
& 82.66 & 92.95 \\

LQ Input
& \underline{4.10} & \underline{32.90}
& {16.72} & {61.38}
& 20.83 & 66.43
& 7.55 & 51.39
& 34.55 & 71.02 \\

\midrule

\textit{Single-View Restoration} \\
\midrule
Restormer~\cite{zamir2022restormer}
& 3.69 & 26.68
& 16.51 & 57.68  
& 21.08 & 67.80 
& \underline{24.94} & 75.12
& \underline{39.20} & \underline{76.12} \\

HI-Diff~\cite{chen2023hierarchical}
& 1.37 & 16.02
& 15.51 & 52.56  
& 10.54 & 51.13
& 7.30 & 52.72
& 33.91 & 71.61 \\

InstructIR~\cite{conde2024high}
& 3.89 & 27.68
& 14.71 & 53.80
& \underline{54.80} & \underline{85.91}
& 18.42 & 74.22
& 32.84 & 68.23 \\

MoCE-IR~\cite{zamfir2024complexityexperts}
& 2.88 & 28.60
& \underline{21.73} & \underline{63.25}
& 45.85 & 84.61
& 16.38 & 65.20
& 38.64 & 72.99 \\

\midrule

\textit{Multi-View Restoration} \\
\midrule

VRT~\cite{liang2024vrt}
& 3.67 & 30.17
& 14.90 & 58.98
& 19.83 & 67.01
& 11.11 & 48.02
& 36.55 & 72.67
\\

FMA-Net~\cite{Youk_2024_CVPR}
&  2.10  &    18.29
&  10.78 &    53.81
&  7.09 &    40.62
&  7.55 &    38.23
&  24.66  &   55.40
\\


VAE$_\text{MVD}$
& 2.84 & 28.70  
& 7.88 & 35.20 
& 8.18 & 60.50
& {13.90} & \underline{76.50}
& 31.20 & 75.00 \\

\textbf{GARD (Ours)}
& \textbf{12.00} & \textbf{67.22}
& \textbf{35.75} & \textbf{74.68} 
& \textbf{62.24} & \textbf{92.37}
& \textbf{35.55} & \textbf{84.73}
& \textbf{56.44} & \textbf{87.45} \\

\bottomrule
\end{tabular}
}
\end{table*}
\begin{figure}[t]
  \centering
  \includegraphics[width=\textwidth]{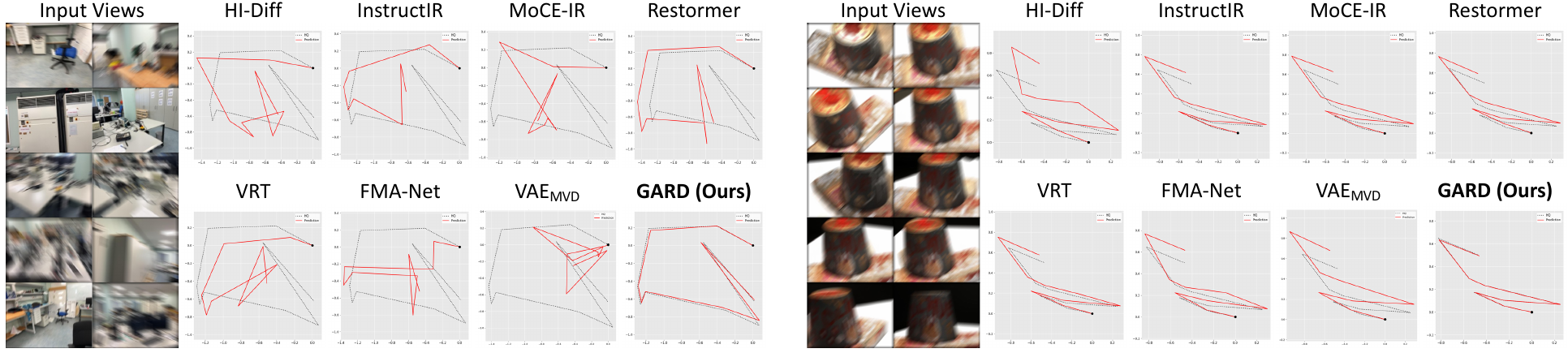}
  \vspace{-15pt}
    \caption{\textbf{Qualitative results for camera trajectory prediction.}
    We visualize the top-down camera trajectories for degraded multi-view inputs. Compared to the baselines, the proposed GARD produces more accurate and geometrically consistent camera pose trajectories. The black dot indicates the starting camera point. Please zoom in for clearer visualization.}
  \label{fig:pose_main}
  \vspace{-10pt}
\end{figure}

\paragraph{Implementation details.}
We employ Depth Anything 3 (DA3)~\cite{depthanything3} as our primary feed-forward 3D reconstruction model, initialized from the DA3-GIANT-1.1~\cite{depthanything3} checkpoint, whose multi-view encoder consists of $L=40$ layers with a hidden dimension of $C=1536$. The proposed GARD denoiser is applied at the $K=18$-th layer of the multi-view encoder. The selected feature levels from the multi-view encoder used by the geometry and RGB decoders are $M=\{20, 28, 34, 40\}$. The GARD denoiser architecture adopts $\text{DiT}^{\text{DH}}$ from RAE~\cite{zheng2025diffusion}, augmented with global attention layers, with encoder and decoder depths of 8 and 6 layers, respectively. The noise perturbation coefficient for training with the interpolated flow matching loss is set to $\alpha=0.3$. The attention alignment layer is set to $J=9$, and the alignment coefficient is set to $\lambda_\text{attn}=1.0$. The target correspondence maps $\mathbf{A}^{*}$ are obtained from point clouds reconstructed by forwarding clean multi-view images through the model. We use $V=10$ multi-view input images for all evaluations. Refer to the supplementary material for further training details, including the datasets, evaluation metrics, and configurations.

\begin{table*}[t]
\caption{\textbf{Quantitative 3D reconstruction results.} We report Overall$\downarrow$ and F-Score$\uparrow$ for 3D reconstruction on the DA3 benchmark~\cite{depthanything3}, comparing single-view and multi-view restoration approaches. The best result is highlighted in \textbf{bold} and the second best is \underline{underlined}.}
\label{tab:recon}
\centering

\resizebox{\linewidth}{!}{
\begin{tabular}{lccccccccc}
\toprule
\multirow{2}{*}{\textbf{Model}}
& \multicolumn{2}{c}{\textbf{HiRoom}}
& \multicolumn{2}{c}{\textbf{ETH3D}}
& \multicolumn{1}{c}{\textbf{DTU}}
& \multicolumn{2}{c}{\textbf{7Scenes}}
& \multicolumn{2}{c}{\textbf{ScanNet$++$}} \\
\cmidrule(lr){2-3} 
\cmidrule(lr){4-5} 
\cmidrule(lr){6-6} 
\cmidrule(lr){7-8} 
\cmidrule(lr){9-10}

& Overall$\downarrow$ & F-score$\uparrow$
& Overall$\downarrow$ & F-score$\uparrow$
& Overall$\downarrow$
& Overall$\downarrow$ & F-score$\uparrow$
& Overall$\downarrow$ & F-score$\uparrow$ \\
\midrule

\textit{Input Conditions} \\
\midrule
HQ Input
& 0.069 & 84.05
& 0.812 & 60.81
& 2.475
& 0.159 & 45.15
& 0.265 & 50.25 \\

LQ Input
& 1.634 & 11.74
& {1.564} & \underline{37.50}
& 6.611
& 0.363 & 18.40
& 0.335 & 24.13 \\

\midrule

\textit{Single-View Restoration} \\
\midrule
Restormer~\cite{zamir2022restormer}
& 0.842 & 11.21
& 2.116 & 33.97 
& 7.272
& 0.388 & 27.92
& {0.326} & \underline{30.45} \\

HI-Diff~\cite{chen2023hierarchical}
& 1.778 & 8.07
& 3.256 & 35.10  
& 7.758
& 0.392 & 25.92
& 0.334 & 25.83 \\

InstructIR~\cite{conde2024high}
& 0.992 & \underline{12.41}
& 2.263 & 33.71  
& \underline{5.563}
& 0.311 & \underline{29.80}
& 0.366 & 26.06 \\

MoCE-IR~\cite{zamfir2024complexityexperts}
&  1.275 & 10.62
&  1.539 & 37.15
&  6.120
&  0.291 & 26.31
&  0.324  & 25.97
\\

\midrule

\textit{Multi-View Restoration} \\
\midrule

VRT~\cite{liang2024vrt}
& 1.289  & 9.45
& \underline{1.493} & 35.14
& 7.570
& 0.538 &  19.53
& \underline{0.319} &  26.72
\\

FMA-Net~\cite{Youk_2024_CVPR}
&  1.988 &   9.61
&  1.865 &   35.65
&  7.415
&  1.368 &   13.37
&  0.389 &   19.66
\\


VAE$_\text{MVD}$
& \underline{0.750} & 11.26
& 2.046 & 25.64   
& 7.745
& \underline{0.259} & 28.16
& 0.343 & 28.38 \\

\textbf{GARD (Ours)}
& \textbf{0.293} & \textbf{18.25}
& \textbf{1.136} & \textbf{45.79} 
& \textbf{4.760}
& \textbf{0.190} & \textbf{36.08}
& \textbf{0.277} & \textbf{35.77} \\


\bottomrule
\end{tabular}
}
\end{table*}
\begin{figure}[t]
  \centering
  \vspace{-15pt}
  \includegraphics[width=\textwidth]{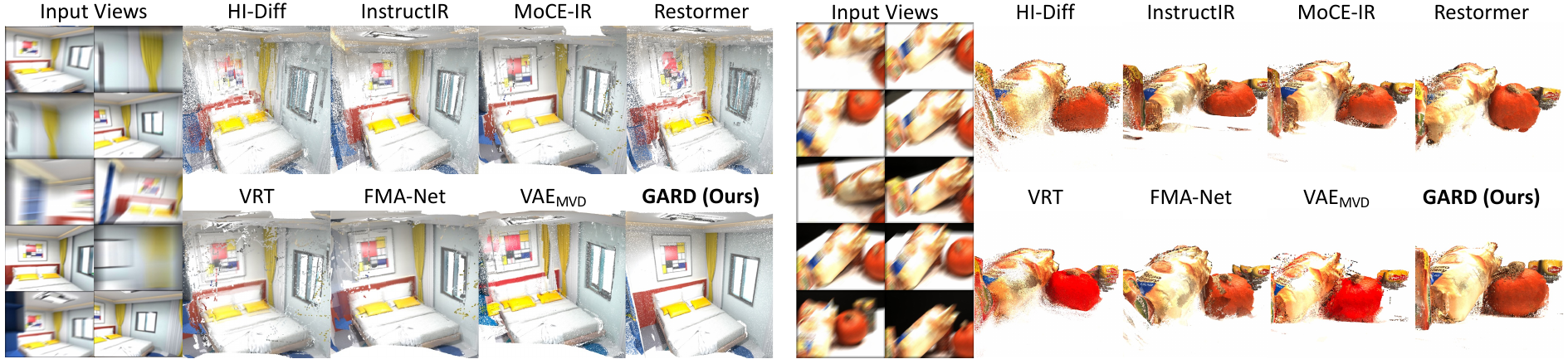}
  \vspace{-15pt}
\caption{\textbf{Qualitative 3D reconstruction results.}
We visualize the reconstructed 3D point clouds from degraded multi-view inputs. Compared with baseline approaches, the proposed GARD produces more accurate and geometrically consistent 3D reconstructions. Please zoom in for clearer visualization.}
  \label{fig:recon_main}
  \vspace{-15pt}
\end{figure}

\subsection{Experimental Results}
\label{sec:exp_results}

\paragraph{Pose estimation.}
Tab.~\ref{tab:pose} and Fig.~\ref{fig:pose_main} report the quantitative and qualitative evaluation results of camera pose estimation (AUC$\uparrow$) across five DA3 benchmarks~\cite{depthanything3} under severe motion blur degradation. Directly applying a feed-forward 3D reconstruction model to degraded inputs results in substantial deterioration in pose estimation accuracy, revealing the limited robustness of feed-forward 3D reconstructors in degraded environments, as these models are predominantly trained and evaluated under clean conditions. Both single-view and multi-view baseline approaches underperform compared with the proposed GARD due to their single-view processing architectures, limited capability to utilize temporally distant frames, and operation within heavily compressed VAE latent spaces during restoration. In contrast, the proposed \ourmodel{} performs denoising directly within the feature space of the feed-forward reconstructor, thereby leveraging geometry-aware feature representations to better preserve structural fidelity and cross-view consistency throughout restoration, which leads to more accurate and stable pose estimation under severe degradation. Refer to Fig.~\ref{fig:fig_suppl_pose} in the supplementary material for additional qualitative visualizations.


\paragraph{3D reconstruction.}
Tab.~\ref{tab:recon} and Fig.~\ref{fig:recon_main} report the quantitative and qualitative evaluation results of 3D reconstruction (Overall$\downarrow$, F-score$\uparrow$) across five DA3 benchmarks~\cite{depthanything3} under severe motion blur degradation. Consistent with the pose estimation results, \ourmodel{} achieves the best performance across all datasets, demonstrating its effectiveness in recovering accurate and complete 3D scene geometry under degraded conditions. While single-view restoration models may improve perceptual image quality, their inability to leverage complementary information across multi-view inputs limits restoration performance and prevents the preservation of geometric consistency, resulting in incomplete or spatially misaligned 3D reconstructions. Furthermore, VAE-based latent restoration approaches suffer from information loss, which limits their ability to recover detailed geometric structures. By operating directly within a geometry-aware feature space, GARD effectively preserves both global structural coherence and fine local geometric details, resulting in more complete and accurate 3D reconstructions. As reconstruction quality depends not only on pose accuracy but also on the preservation of detailed geometric structures such as depth maps, we additionally provide depth evaluation results in Tab.~\ref{supple_tab:da3_giant_depth} and Fig.~\ref{fig:fig_suppl_depth} of the supplementary material. Additional qualitative reconstruction visualizations are presented in Fig.~\ref{fig:fig_supple_recon} of the supplementary material.

\begin{figure}[t]
  \centering
  \includegraphics[width=\textwidth]{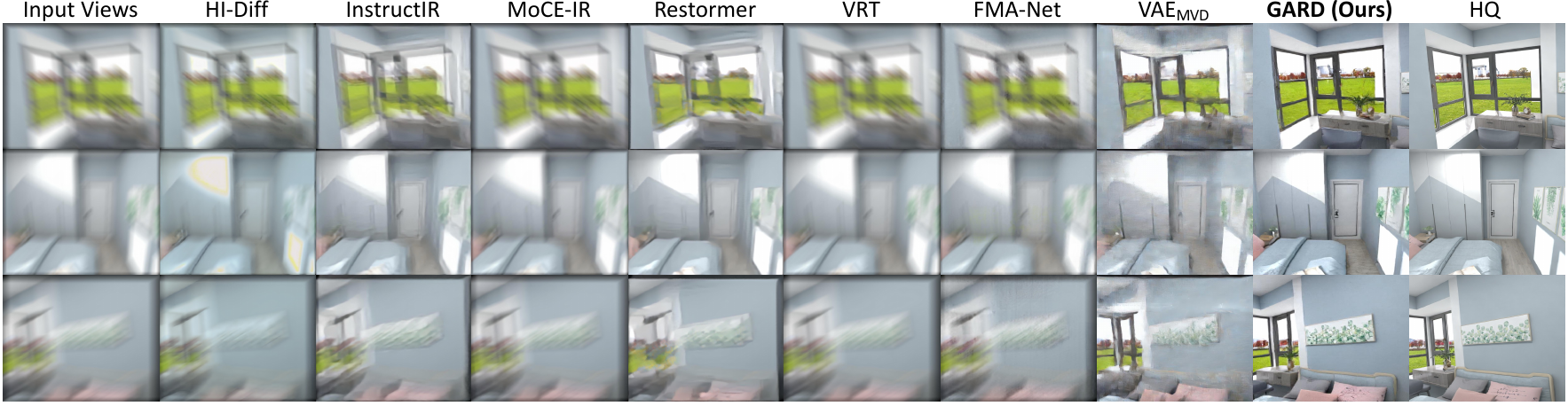}
  \vspace{-15pt}
\caption{\textbf{Qualitative image restoration results.}
We visualize restored RGB images from degraded multi-view inputs. Compared with baseline approaches, the proposed GARD effectively recovers high-fidelity multi-view images while preserving fine-grained details. Please zoom in for clearer visualization.}
  \label{fig:rgb_main}
\end{figure}

\paragraph{Image restoration.}
Tab.~\ref{tab:image_metric} and Fig.~\ref{fig:rgb_main} report the quantitative and qualitative evaluation results of image restoration quality (PSNR$\uparrow$, LPIPS$\downarrow$) across five DA3 benchmarks~\cite{depthanything3} under severe motion blur degradation. Since single-view restoration models are unable to exploit complementary information across multi-view inputs, they exhibit limited restoration capability under severe degradations. For multi-view restoration approaches, although video restoration models such as VRT~\cite{liang2024vrt} and FMA-Net~\cite{Youk_2024_CVPR} can leverage multi-view information, long-range temporal input frames substantially challenge effective restoration, as these models are primarily designed for temporally adjacent frame sequences. Among the compared baselines, only the trained VAE$_\text{MVD}$ achieves noticeable performance improvements over preceding approaches. Nevertheless, the proposed GARD attains the best overall performance, achieving higher PSNR and lower LPIPS than both single-view and VAE-based multi-view restoration baselines, demonstrating that performing restoration denoising in a high-dimensional geometry-aware feature space more effectively preserves visual fidelity and structural consistency. Additional qualitative comparisons are provided in Fig.~\ref{fig:fig_suppl_rgb_recon} of the supplementary material.

\subsection{Ablation Experiments}

\paragraph{GARD denoiser training components.}
Tab.~\ref{tab:abl_gard_training} presents an ablation study on the training components of the GARD denoiser, namely the interpolated flow matching loss and the attention alignment loss. Models A--D correspond to different configurations trained with or without these components, where Model D denotes the full configuration used to train the proposed GARD denoiser. The results show that attention alignment does not consistently improve performance when used with the standard flow matching objective. However, when combined with interpolated flow matching, it yields consistent performance gains. This can be attributed to the fact that the standard flow matching objective initializes the denoising process from a pure Gaussian noise distribution, which contains no structural prior, making correspondence learning difficult even with attention supervision. In contrast, interpolated flow matching initializes the denoising process from a mixture of Gaussian noise and the degraded low-quality (LQ) input distribution, thereby introducing partial structural information. This structural prior enables the GARD denoiser to better learn the mapping from the low-quality (LQ) degraded distribution to the high-quality (HQ) clean distribution, leveraging attention alignment signals and ultimately improving restoration performance. The effect of attention alignment is visualized in Fig.~\ref{fig:suppl_attn} of the supplementary material.

\paragraph{Number of input views.}
Tab.~\ref{tab:abl_maxview} presents an ablation study examining the impact of the number of input views used for multi-view restoration. Specifically, we investigate how increasing the number of views affects both camera pose estimation accuracy and 3D reconstruction quality across the DA3 benchmarks. Incorporating additional input views consistently yields improvements in both pose estimation and reconstruction performance, demonstrating that multi-view observations provide complementary geometric and visual information beneficial for multi-view restoration. We note that, in the HiRoom benchmark within DA3, each scene contains at most 20 views, accordingly, results are reported only up to 10 input views.


\begin{table*}[t]
\caption{\textbf{Quantitative restoration results.} We report PSNR$\uparrow$ and LPIPS$\downarrow$ for image restoration on the DA3 benchmark~\cite{depthanything3}, comparing single-view and multi-view restoration approaches. The best result is highlighted in \textbf{bold} and the second best is \underline{underlined}.}
\label{tab:image_metric}
\centering

\resizebox{\linewidth}{!}{
\begin{tabular}{lccccc ccccc ccccc ccccc}
\toprule
\multirow{2}{*}{\textbf{Model}}
& \multicolumn{2}{c}{\textbf{HiRoom}}
& \multicolumn{2}{c}{\textbf{ETH3D}}
& \multicolumn{2}{c}{\textbf{DTU}}
& \multicolumn{2}{c}{\textbf{7Scenes}}
& \multicolumn{2}{c}{\textbf{ScanNet$++$}} \\
\cmidrule(lr){2-3} \cmidrule(lr){4-5} \cmidrule(lr){6-7} \cmidrule(lr){8-9} \cmidrule(lr){10-11}

& PSNR$\uparrow$ & LPIPS$\downarrow$
& PSNR$\uparrow$ & LPIPS$\downarrow$
& PSNR$\uparrow$ & LPIPS$\downarrow$
& PSNR$\uparrow$ & LPIPS$\downarrow$
& PSNR$\uparrow$ & LPIPS$\downarrow$ \\



\midrule

\textit{Single-View Restoration} \\
\midrule
Restormer~\cite{zamir2022restormer}
& 17.49 & 0.544
& 20.97	& 0.672
& 17.73 & 0.588
& 21.30 & {0.428}
& 21.50 & 0.415  \\

HI-Diff~\cite{chen2023hierarchical}
& 17.35 & 0.551
& 20.45 & \underline{0.586}
& 17.39  & 0.591
& 19.82	& {0.428}
& 20.68 & 0.413  \\

InstructIR~\cite{conde2024high}
& 17.51 & 0.549
& 20.93	& 0.666
& 20.38 & 0.558
& 20.93 & 0.440
& 21.15 & 0.402  \\

MoCE-IR~\cite{zamfir2024complexityexperts}
& 17.69  &    0.560
& 21.00  &    0.611
&  20.38 &    0.562
&  20.60 &    0.491
&  \underline{21.19} &    0.431
\\

\midrule
\textit{Multi-View Restoration} \\
\midrule

VRT~\cite{liang2024vrt}
& 17.47  &    0.562
& 20.82  &    0.606
& 17.61  &    0.606
& 19.79  &    0.510
& 20.83  &     0.437
\\

FMA-Net~\cite{Youk_2024_CVPR}
& 17.14 &   0.551
& 20.65 &   \textbf{0.571}
& 17.13 & 0.572
& 18.84 & 0.534
& 19.98 & 0.455
\\


VAE$_\text{MVD}$
&    \underline{19.76}	&   \underline{0.493}
&     \underline{21.37}	&   0.638
&    \underline{20.54}	&  \underline{0.434}
&     \underline{21.74}	&   \underline{0.404}
&    \underline{21.19}	&  \underline{0.379} \\

\textbf{GARD (Ours)}
& \textbf{21.89}	& \textbf{0.362}
& \textbf{21.88}	& {0.635}
& \textbf{21.25} & \textbf{0.418}
& \textbf{22.67} &	\textbf{0.249}
& \textbf{22.19} & \textbf{0.345} \\

\bottomrule
\end{tabular}
}
\end{table*}

\begin{table}[h]
\centering

\subfloat[
\textbf{Pose estimation accuracy.} We report the area under the curve (AUC30)$\uparrow$ for pose evaluation.
\label{tab:lr_ablation_pose}
]{
\begin{minipage}{0.48\linewidth}
\centering
\resizebox{\textwidth}{!}{
\begin{tabular}{lccccc}
\toprule
\multirow{2}{*}{Model} &
\multirow{2}{*}{Interp. Flow} &
\multirow{2}{*}{Alignment} &
\multicolumn{3}{c}{AUC30$\uparrow$} \\
\cmidrule(lr){4-6}
& & & ETH3D & DTU & ScanNet++ \\
\midrule
A    & \texttimes & \texttimes & 67.30 & 87.21 & 84.12 \\
B    & \texttimes & \checkmark & 66.42 & 85.49 & 84.90 \\
C    & \checkmark & \texttimes & \underline{73.85} & \underline{89.99} & \underline{85.90} \\
\midrule
D (Full) & \checkmark & \checkmark & \textbf{74.68} & \textbf{92.37} & \textbf{87.45} \\
\bottomrule
\end{tabular}}
\end{minipage}
}
\hfill
\subfloat[
\textbf{3D reconstruction accuracy.} We report Overall$\downarrow$ for DTU, while F-score$\uparrow$ is reported for the remaining benchmarks.
\label{tab:lr_ablation_recon}
]{
\begin{minipage}{0.48\linewidth}
\centering
\resizebox{\textwidth}{!}{
\begin{tabular}{lccccc}
\toprule
\multirow{2}{*}{Model} &
\multirow{2}{*}{Interp. Flow} &
\multirow{2}{*}{Alignment} &
\multicolumn{3}{c}{F-score$\uparrow$ / Overall$\downarrow$} \\
\cmidrule(lr){4-6}
& & & ETH3D & DTU & ScanNet++ \\
\midrule
A   & \texttimes & \texttimes & 39.91 & 5.43 & 31.52 \\
B    & \texttimes & \checkmark & 38.44 & 5.40 & 30.63 \\
C    & \checkmark & \texttimes & \underline{44.65} & \underline{4.92} & \underline{32.40} \\
\midrule
D (Full) & \checkmark & \checkmark & \textbf{45.79} & \textbf{4.76} & \textbf{35.77} \\
\bottomrule
\end{tabular}}
\end{minipage}
}
\vspace{-5pt}
\caption{\textbf{Ablation on GARD training components.}
We conduct an ablation study on the training components of GARD across three representative benchmarks spanning outdoor, object-centric, and indoor scenes: ETH3D~\cite{schops2017multi}, DTU~\cite{aanaes2016large}, and ScanNet++~\cite{yeshwanth2023scannet++}, respectively.}
\label{tab:abl_gard_training}\vspace{-10pt}
\end{table}

\begin{table*}[h]
\centering
\subfloat[
\textbf{Pose estimation accuracy.} We report the area under the curve (AUC30)$\uparrow$ for pose evaluation.
\label{tab:view_ablation_a}
]{
\begin{minipage}{0.48\linewidth}
\centering
\resizebox{\textwidth}{!}{
\begin{tabular}{lccccc}
\toprule
\multirow{2}{*}{\textbf{\# Views}} &
\multicolumn{5}{c}{AUC30$\uparrow$} \\
\cmidrule(lr){2-6}
& HiRoom & ETH3D & DTU & 7Scenes & ScanNet$++$ \\
\midrule
4 views  & \underline{63.61} & 35.80 & 84.25 & 71.62 & 60.12 \\
10 views & \textbf{67.22} & 74.68 & 92.37 & 84.73 & 87.45 \\
30 views & - & \underline{82.35} & \underline{94.29}  & \underline{85.22} & \underline{94.61} \\
50 views & - & \textbf{82.62} & \textbf{98.00} & \textbf{86.85} &  \textbf{95.30} \\
\bottomrule
\end{tabular}}
\end{minipage}
}
\hfill
\subfloat[
\textbf{3D reconstruction accuracy.} We report Overall$\downarrow$ for DTU, while F-score$\uparrow$ is reported for the remaining benchmarks.
\label{tab:view_ablation_b}
]{
\begin{minipage}{0.48\linewidth}
\centering
\resizebox{\textwidth}{!}{
\begin{tabular}{lccccc}
\toprule
\multirow{2}{*}{\textbf{\# Views}} &
\multicolumn{5}{c}{F-score$\uparrow$ / Overall$\downarrow$} \\
\cmidrule(lr){2-6}
& HiRoom & ETH3D & DTU & 7Scenes & ScanNet$++$ \\
\midrule
4 views  & \underline{8.65} & 10.56 & 6.95 & 12.00 & 10.10 \\
10 views & \textbf{18.25} & 45.79 & 4.76 & 36.08 & 35.77 \\
30 views & - & \underline{58.82} & \underline{3.58} & \underline{40.47} & \underline{57.62} \\
50 views & - & \textbf{65.46} & \textbf{2.03} & \textbf{45.92} &  \textbf{65.72} \\
\bottomrule
\end{tabular}}
\end{minipage}
}
\vspace{-10pt}
\caption{\textbf{Ablation study on the number of input views.}
Increasing the number of views improves both camera pose estimation accuracy and 3D reconstruction quality, indicating that richer cross-view information substantially benefits geometric reconstruction.}
\label{tab:abl_maxview}
\vspace{-10pt}
\end{table*}

\section{Conclusion}
\label{sec:conclusion}

We presented \textbf{GARD}, a Geometry-Aware Restoration Denoising framework that performs image restoration in the geometry-aware feature space of feed-forward 3D reconstruction models. By leveraging geometry-aware representations, GARD effectively refines degraded features, which are subsequently passed to both the 3D geometry decoder and the RGB image decoder to recover accurate 3D scene geometry and high-quality restored multi-view RGB images. This representation-level denoising strategy enables the joint restoration of scene structure and appearance from degraded multi-view inputs. Extensive experiments demonstrate that GARD consistently outperforms existing single-view and multi-view restoration methods, highlighting the effectiveness of restoration in geometry-aware feature space.

\paragraph{Limitation and future directions.}
As a diffusion-based method, GARD requires iterative denoising steps, limiting its efficiency in latency-sensitive settings. Future work will focus on more efficient denoiser designs and multi-layer denoising strategies to further reduce error accumulation and improve reconstruction quality.

\clearpage

\bibliographystyle{plain}
\bibliography{neurips_2026}





\clearpage
\appendix
\setcounter{page}{1}
\section*{\Large Appendix}



\section{Extended Experimental Results}

\subsection{Feature Similarity Analysis}
\begin{figure}[h]
    \centering
    \includegraphics[width=\linewidth]{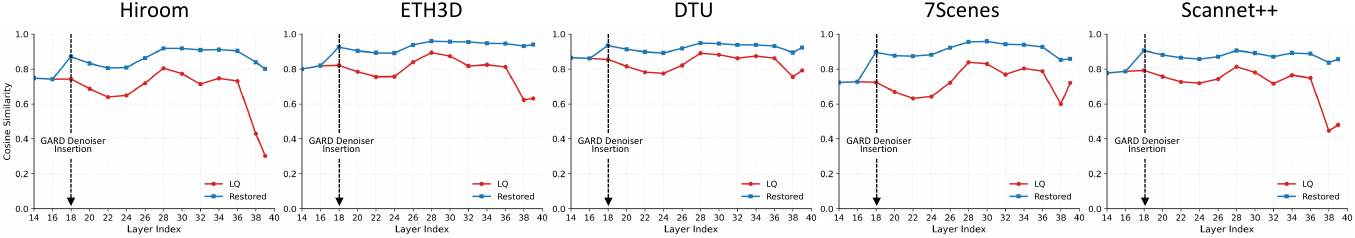}
    \caption{
    \textbf{Feature similarity analysis across layers.}
    We evaluate the cosine similarity between the restored feature representations and the corresponding clean HQ representations across the multi-view encoder layers of the feed-forward 3D reconstruction model~\cite{depthanything3}. The GARD denoiser is applied at layer $K=18$. Across all DA3 benchmark~\cite{depthanything3} datasets, the feature similarity of the degraded LQ representations (red) progressively decreases in deeper layers due to the accumulation of degradation-induced feature distortion. In contrast, the restored representations produced by the GARD denoiser (blue) recover feature representations that remain substantially closer to the clean HQ representations throughout the deeper layers. This demonstrates that the GARD denoiser effectively restores geometry-aware features and suppresses degradation propagation, resulting in substantially improved consistency with the clean HQ representations. Please zoom in for clearer visualization.
    }
    \label{fig:suppl_feat_sim}
\end{figure}

To further validate the effectiveness of the proposed GARD framework, we investigate the feature similarity across the transformer layers of the feed-forward reconstructor. Since the GARD denoiser is trained to map low-quality (LQ) degraded feature representations to their corresponding high-quality (HQ) clean representations, we measure the similarity between the restored representations and the clean representations to assess the extent to which the degraded feature representations are effectively recovered. Specifically, the feature similarity at transformer layer $l$ is computed as follows:
\begin{equation}
\mathrm{Sim}^{l} = 
\frac{
\mathbf{z}_{\text{res}}^{l}
\cdot 
\mathbf{z}_{\text{clean}}^{l}
}{
\left\lVert \mathbf{z}_{\text{res}}^{l} \right\rVert
\left\lVert \mathbf{z}_{\text{clean}}^{l} \right\rVert
},
\end{equation}
where $\mathbf{z}_{\text{res}}^{l}$ and $\mathbf{z}_{\text{clean}}^{l}$ are the restored and clean multi-view representations at layer $l$, respectively. Fig.~\ref{fig:suppl_feat_sim} visualizes this similarity for two independent scenes. In the baseline setting, without using the GARD denoiser, the similarity to the clean representation progressively decreases as the representations propagate through deeper layers, indicating that severe degradations cause a gradual loss of geometric information in the encoded features. In contrast, applying GARD at an early layer initially improves the similarity between restored and clean representations. These results provide a rationale for the improved reconstruction and pose estimation performance, demonstrating that GARD effectively maintains high-fidelity representations throughout the pipeline.

\subsection{Feature Cost Volume Visualization}
\begin{figure}[h]
    \centering
    \includegraphics[width=0.95\linewidth]{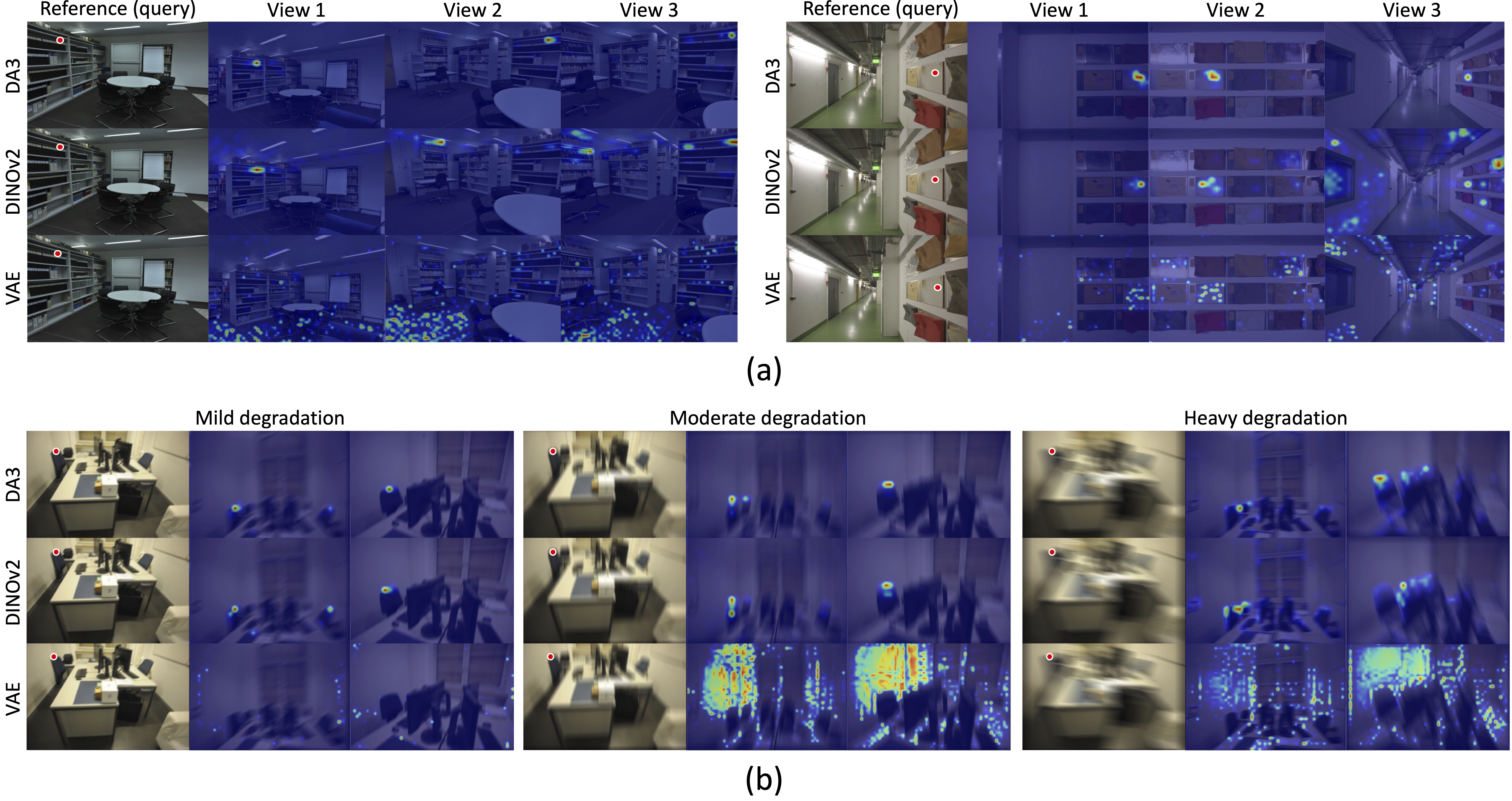}
    \caption{
    \textbf{Correspondence visualization of feature cost volumes.}
    Cross-view correspondence visualization of feature cost volumes constructed from VAE~\cite{kingma2013auto}, DINOv2~\cite{oquab2023dinov2}, and DA3~\cite{depthanything3} feature cost volumes. Please zoom in for clearer visualization.
    }
    \label{fig:suppl_cost_volume}
\end{figure}

Fig.~\ref{fig:suppl_cost_volume} further presents qualitative correspondence attention visualizations of the three feature cost volumes~\cite{kingma2013auto, oquab2023dinov2, depthanything3} analyzed in Fig.~\ref{fig:geo_aware} of the main paper. Specifically, for each red query point in the reference view, we visualize the attention region under (a) clean high-quality (HQ) multi-view inputs and (b) progressively increasing degradation levels of degraded multi-view inputs (mild, moderate, and heavy). Both quantitative and qualitative results reveal a clear performance hierarchy of DA3 $>$ DINOv2 $\gg$ VAE in terms of correspondence accuracy and robustness. Comparing the three feature cost volume representations through attention correspondence, DA3 features produce the sharpest and most geometrically consistent correspondences across views under both clean and degraded conditions, whereas DINOv2 achieves reasonably accurate matching with less precise and less stable response regions. In contrast, VAE features exhibit highly scattered and ambiguous correspondences under both clean and degraded conditions, reflecting limited geometry-aware representation quality caused by heavy latent compression. As the degradation severity increases, DA3 maintains strong localization and geometric consistency, whereas DINOv2 gradually deteriorates and VAE fails to preserve reliable correspondences.

\subsection{Multi-View Depth Estimation}
Tab.~\ref{supple_tab:da3_giant_depth} further presents the multi-view depth estimation evaluation. We report AbsRel$\downarrow$ and $\delta_1\uparrow$, which measure the absolute relative depth error and the percentage of predicted depths within a threshold ratio of $1.25$ from the ground-truth depth, respectively. GARD demonstrates superior depth estimation performance on the DA3~\cite{depthanything3} benchmark datasets, outperforming all baseline methods on four datasets while exhibiting only a marginal performance gap on DTU. These results underscore the effectiveness of restoration in a geometry-aware representation space for enhancing downstream multi-view depth estimation. By restoring degraded observations directly in this space, our method preserves cross-view geometric consistency and fine-grained structural details that are essential for accurate depth prediction. In contrast, single-view restoration methods are inherently unable to enforce geometric consistency across views, while VAE-based multi-view baselines are constrained by compression bottlenecks that limit the preservation of detailed geometric structures. As a result, baseline methods are more vulnerable to severe degradations, often producing blurred structures and visible artifacts, whereas our method yields sharper and more geometrically coherent depth maps across views, as further illustrated in the qualitative depth evaluation results in Fig.~\ref{fig:fig_suppl_depth}.
\begin{table*}[h]
\caption{\textbf{Quantitative depth estimation results.} We report AbsRel$\downarrow$ and $\delta_1\uparrow$ across the DA3 benchmarks~\cite{depthanything3}. The best result is highlighted in \textbf{bold} and the second best is \underline{underlined}.}
\label{supple_tab:da3_giant_depth}
\centering

\resizebox{\linewidth}{!}{
\begin{tabular}{lccccc ccccc ccccc ccccc}
\toprule
\multirow{2}{*}{\textbf{Model}}
& \multicolumn{2}{c}{\textbf{HiRoom}}
& \multicolumn{2}{c}{\textbf{ETH3D}}
& \multicolumn{2}{c}{\textbf{DTU}}
& \multicolumn{2}{c}{\textbf{7Scenes}}
& \multicolumn{2}{c}{\textbf{ScanNet$++$}} \\
\cmidrule(lr){2-3} \cmidrule(lr){4-5} \cmidrule(lr){6-7} \cmidrule(lr){8-9} \cmidrule(lr){10-11}

& AbsRel$\downarrow$ & $\delta_1\uparrow$
& AbsRel$\downarrow$ & $\delta_1\uparrow$
& AbsRel$\downarrow$ & $\delta_1\uparrow$
& AbsRel$\downarrow$ & $\delta_1\uparrow$
& AbsRel$\downarrow$ & $\delta_1\uparrow$ \\
\midrule

\textit{Input Conditions} \\
\midrule
HQ Input
& 0.012 & 99.4
& 0.021 & 99.7
& 0.055 & 95.2
& 0.062 & 93.7
& 0.035 & 97.5 \\

LQ Input
& 0.191 & 78.6
& 0.087 & 90.4
& 0.061 & 94.5
& 0.137 & 81.9
& 0.092 & 89.5 \\

\midrule

\textit{Single-View Restoration} \\
\midrule

Restormer~\cite{zamir2022restormer}
& 0.210 & 74.1
& 0.140 & 82.2 
& 0.061 & 94.6
& 0.117 & 85.0
& 0.097 & 89.1 \\

HI-Diff~\cite{chen2023hierarchical}
& 0.239 & 70.2
& 0.105 & 86.3  
& 0.065 & 93.8
& 0.131 & 82.2
& 0.097 &  89.2 \\

InstructIR~\cite{conde2024high}
& 0.227 & 71.7
& 0.142 & 81.3 
& \underline{0.056} & 95.0
& 0.124 & 83.3
& 0.128 &  83.4 \\

MoCE-IR~\cite{zamfir2024complexityexperts}
& 0.198  &   76.8
& 0.096  &   88.3
& \textbf{0.055}  &   95.6
& 0.125  &   84.5
& 0.092  &   89.6
\\

\midrule

\textit{Multi-View Restoration} \\
\midrule

VRT~\cite{liang2024vrt}
& 0.193  &   77.9
& 0.087  &   90.2
& 0.060  &   94.7
& 0.144  &   80.5
& 0.092  &   89.8
\\

FMA-Net~\cite{Youk_2024_CVPR}
& 0.240 &    70.8
& 0.101 &    88.0
& 0.064 &    94.1
& 0.153 &    76.8
& 0.121 &    85.5
\\

VAE$_\text{MVD}$
& \underline{0.188} & \underline{80.2}
& \underline{0.069} & \underline{95.4}
& {0.057} & \underline{95.9}
& \underline{0.093} & \underline{89.0}
& \underline{0.068} & \underline{93.2} \\

\textbf{GARD (Ours)}
& \textbf{0.048} & \textbf{97.2}
& \textbf{0.036} & \textbf{98.4}
& {0.057} & \textbf{96.1}
& \textbf{0.071} & \textbf{92.7}
& \textbf{0.040} & \textbf{96.7}  \\


\bottomrule
\end{tabular}
}
\end{table*}

\subsection{Multi-View Restoration}
\begin{figure}[h]
    \centering
   \includegraphics[width=0.95\linewidth]{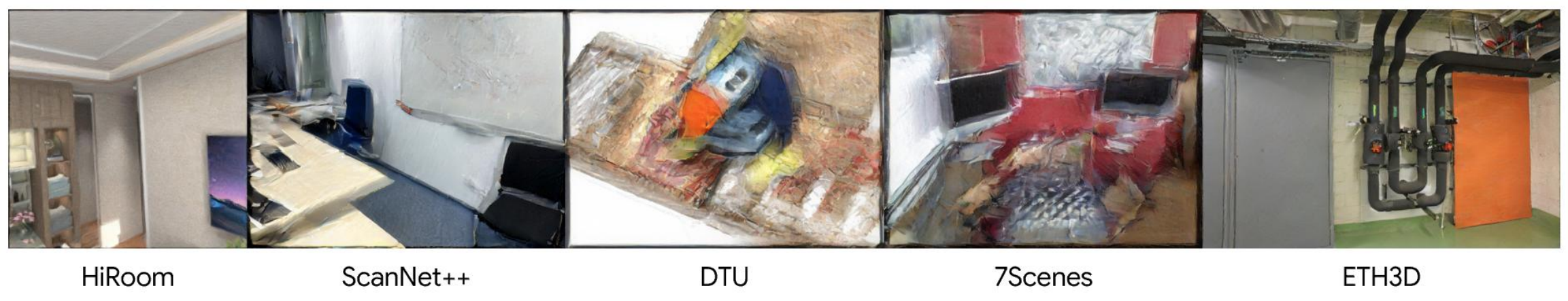}
    \vspace{-5pt}
    \caption{\textbf{Multi-view image restoration results of SIR-Diff~\cite{mao2025sir}.} We show qualitative results on the DA3 benchmark~\cite{depthanything3} based on our reproduction of SIR-Diff.}
    \label{fig:fig_suppl_sirdiff}
\end{figure}

We additionally compare our method with SIR-Diff~\cite{mao2025sir}, a multi-view diffusion-based restoration model that leverages Spatial-3D ResNet blocks and 3D self-attention Transformers for effective restoration of degraded multi-view inputs. As the official pretrained checkpoints and training code were not publicly available, we contacted the authors for clarification and implemented the method based on the available unofficial training code together with the details provided in the paper. While we carefully followed the described training protocol to ensure a fair comparison, some components of the unofficial implementation were incomplete, which made full reproduction of the reported results challenging. Nevertheless, we made our best effort to faithfully follow the method as described. The quantitative comparisons are presented in Tab.~\ref{tab:sirdiff_compare}, and the qualitative results are provided in Fig.~\ref{fig:fig_suppl_sirdiff}. While SIR-Diff can produce reasonable restorations under mild degradation settings, its performance degrades noticeably under stronger corruption levels, where it often introduces artifacts and structural inconsistencies. As a result, its effectiveness for downstream geometric tasks becomes limited in challenging scenarios.
\vspace{-10pt}
\begin{table*}[h]
\centering
\subfloat[
\textbf{Pose estimation accuracy.} We report the area under the curve (AUC30)$\uparrow$ for pose evaluation.
\label{tab:sirdiff_pose}
]{
\begin{minipage}{0.48\linewidth}
\centering
\resizebox{\textwidth}{!}{
\begin{tabular}{lccccc}
\toprule
\multirow{2}{*}{\textbf{Model}} &
\multicolumn{5}{c}{AUC30$\uparrow$} \\
\cmidrule(lr){2-6}
& HiRoom & ETH3D & DTU & 7Scenes & ScanNet$++$ \\
\midrule
SIR-Diff~\cite{mao2025sir} & 9.73 & 20.94 & 16.59 & 2.79 & 30.02 \\
\textbf{GARD (Ours)} & \textbf{67.22} & \textbf{74.68} & \textbf{92.37} & \textbf{84.73} & \textbf{87.45} \\
\bottomrule
\end{tabular}}
\end{minipage}
}
\hfill
\subfloat[
\textbf{3D reconstruction accuracy.} We report Overall$\downarrow$ for DTU, while F-score$\uparrow$ is reported for the remaining benchmarks.
\label{tab:sirdiff_recon}
]{
\begin{minipage}{0.48\linewidth}
\centering
\resizebox{\textwidth}{!}{
\begin{tabular}{lccccc}
\toprule
\multirow{2}{*}{\textbf{Model}} &
\multicolumn{5}{c}{F-score$\uparrow$ / Overall$\downarrow$} \\
\cmidrule(lr){2-6}
& HiRoom & ETH3D & DTU & 7Scenes & ScanNet$++$ \\
\midrule
SIR-Diff~\cite{mao2025sir}  & 5.12 & 11.72 & 8.101 & 0.00 & 12.74 \\
\textbf{GARD (Ours)} & \textbf{18.25} & \textbf{45.79} & \textbf{4.760} & \textbf{36.08} & \textbf{35.77} \\
\bottomrule
\end{tabular}}
\end{minipage}
}
\vspace{-10pt}
\caption{\textbf{Quantitative evaluation of SIR-Diff~\cite{mao2025sir}.} We report the pose estimation and 3D reconstruction results on the DA3 benchmark~\cite{depthanything3}, comparing SIR-Diff~\cite{mao2025sir} with our GARD framework.}
\label{tab:sirdiff_compare}
\end{table*}

\vspace{-15pt}


\subsection{Multi-View Encoder Backbone}
We further evaluate the effectiveness of the GARD denoiser across encoder backbones of varying capacity by training it on a different multi-view encoder. Specifically, we train the GARD denoiser using representations from DA3-BASE. In contrast to DA3-GIANT, which comprises 40 transformer layers and 1.10B parameters, DA3-BASE consists of 12 layers and 0.11B parameters. Following the same design as DA3-GIANT, we apply the GARD denoiser within the DA3-BASE feature space by inserting it at encoder layer $K=4$, denoted as \textbf{GARD}$_{\textbf{BASE}}$. We report the pose estimation, 3D reconstruction, and depth estimation results in Tab.~\ref{tab:suppl_tab:da3_base_pose}, Tab.~\ref{supple_tab:da3_base_recon}, and Tab.~\ref{supple_tab:da3_base_depth}, respectively.
\begin{table*}[h]
\caption{\textbf{Quantitative pose estimation results.} We report AUC5$\uparrow$ and AUC30$\uparrow$ for camera pose estimation on the DA3 benchmark~\cite{depthanything3} using the DA3-BASE~\cite{depthanything3} checkpoint as the feed-forward 3D reconstruction model. The best result is highlighted in \textbf{bold} and the second best is \underline{underlined}.}
\label{tab:suppl_tab:da3_base_pose}
\centering

\resizebox{\linewidth}{!}{
\begin{tabular}{lcccccccccccccccccccc}
\toprule
\multirow{2}{*}{\textbf{Model}}
& \multicolumn{2}{c}{\textbf{HiRoom}}
& \multicolumn{2}{c}{\textbf{ETH3D}}
& \multicolumn{2}{c}{\textbf{DTU}}
& \multicolumn{2}{c}{\textbf{7Scenes}}
& \multicolumn{2}{c}{\textbf{ScanNet$++$}} \\
\cmidrule(lr){2-3} \cmidrule(lr){4-5} \cmidrule(lr){6-7} \cmidrule(lr){8-9} \cmidrule(lr){10-11}

& AUC5$\uparrow$ & AUC30$\uparrow$
& AUC5$\uparrow$ & AUC30$\uparrow$
& AUC5$\uparrow$ & AUC30$\uparrow$
& AUC5$\uparrow$ & AUC30$\uparrow$
& AUC5$\uparrow$ & AUC30$\uparrow$ \\
\midrule

\textit{Input Conditions} \\
\midrule

HQ Input
&   35.21        &       84.11
&   23.79         &      62.61
&   53.53        &       91.21
&   34.47        &       83.02
&   37.68        &       69.81
\\


LQ Input
&   0.56    &   12.14   
&   1.01    &   11.44    
&   5.97    &   58.78    
&   3.04    &   28.26    
&   0.40    &   10.49
\\

\midrule

\textit{Single-View Restoration} \\
\midrule

Restormer~\cite{zamir2022restormer}
&   0.45    &       12.78
&   \underline{2.18}    &       17.25
&   4.44    &       56.56
&   4.88    &       48.20
&   \underline{5.20}    &       \underline{37.84}
\\


HI-Diff~\cite{chen2023hierarchical}
&   0.19    &       05.68
&   1.13    &       19.33
&   4.84    &       50.57
&   1.96    &       25.55
&   3.02    &       26.56
\\


InstructIR~\cite{conde2024high}
&   0.66    &       11.62
&   1.41    &       16.58
&   \underline{22.76}    &       \underline{76.01}
&   3.61    &        39.58
&   2.75    &       25.80
\\

MoCE-IR~\cite{zamfir2024complexityexperts}
&   0.91    &     17.15
&   2.02   &      \underline{22.36}
&   16.92    &     67.94  
&   4.12    &      35.75 
&   4.24    &     27.72   
\\

\midrule

\textit{Multi-View Restoration} \\
\midrule

VRT~\cite{liang2024vrt}
&    0.62   &    12.79   
&    1.65   &    16.70   
&    6.38   &    57.11   
&    1.33   &    24.21   
&    2.40   &    23.30
\\

FMA-Net~\cite{Youk_2024_CVPR}
&   0.04    &       06.92
&   0.80    &       15.35
&   2.52    &       36.87
&   0.25   &        12.96
&   1.00    &       19.15
\\

VAE$_\text{MVD}$
&    \underline{7.88}   &     \textbf{58.91  }
&    0.24   &     7.05 
&    11.11   &    60.42  
&    \underline{11.49}   &    \underline{51.48}
&    1.37   &     20.60
\\


\textbf{GARD$_\textbf{BASE}$ (Ours)}
&    \textbf{8.14}   &       \underline{58.60}
&    \textbf{6.50}   &       \textbf{32.55}
&    \textbf{24.06}   &      \textbf{76.12}
&    \textbf{23.11}   &      \textbf{68.56}
&    \textbf{17.22}   &      \textbf{55.69}
\\

\bottomrule
\end{tabular}
}
\end{table*}

\begin{table*}[h]
\caption{\textbf{Quantitative 3D reconstruction results.} We report Overall$\downarrow$ and F-Score$\uparrow$ for 3D reconstruction on the DA3 benchmark~\cite{depthanything3} using the DA3-BASE~\cite{depthanything3} checkpoint as the feed-forward 3D reconstruction model. The best result is highlighted in \textbf{bold} and the second best is \underline{underlined}.}
\label{supple_tab:da3_base_recon}
\centering

\resizebox{\linewidth}{!}{
\begin{tabular}{lccccccccc}
\toprule
\multirow{2}{*}{\textbf{Model}}
& \multicolumn{2}{c}{\textbf{HiRoom}}
& \multicolumn{2}{c}{\textbf{ETH3D}}
& \multicolumn{1}{c}{\textbf{DTU}}
& \multicolumn{2}{c}{\textbf{7Scenes}}
& \multicolumn{2}{c}{\textbf{ScanNet$++$}} \\
\cmidrule(lr){2-3}
\cmidrule(lr){4-5}
\cmidrule(lr){6-6}
\cmidrule(lr){7-8}
\cmidrule(lr){9-10}

& Overall$\downarrow$ & F-score$\uparrow$
& Overall$\downarrow$ & F-score$\uparrow$
& Overall$\downarrow$
& Overall$\downarrow$ & F-score$\uparrow$
& Overall$\downarrow$ & F-score$\uparrow$ \\
\midrule

\multicolumn{10}{l}{\textit{Input Conditions}} \\
\midrule


HQ Input
&  0.349     &      22.29
&  1.220     &      37.92
&  5.843
&  0.186     &      31.89
&  0.340    &       26.09
\\


LQ Input
&  0.946     &      7.58
&  3.337     &      11.85
&  8.429
&  0.529     &      12.56
&  \underline{0.476}    &       \underline{13.12}
\\

\midrule

\multicolumn{10}{l}{\textit{Single-View Restoration}} \\
\midrule


Restormer~\cite{zamir2022restormer}
&   0.889    &       7.67
&   3.212    &       \underline{14.67}
&   9.064
&   0.755   &       13.24
&   0.489   &       12.16
\\


HI-Diff~\cite{chen2023hierarchical}
&   1.173    &       4.58
&   3.737     &      8.02
&   8.826
&   0.551   &       12.95
&   0.612    &      8.33
\\


InstructIR~\cite{conde2024high}
&  1.183     &     6.23
&  3.653    &      7.29
&  7.960
&  0.501     &     14.19
&  0.562    &      9.03
\\

MoCE-IR~\cite{zamfir2024complexityexperts}
&   0.870   &      7.81
&   \underline{2.909}    &     13.95
&   7.942   
&   0.504    &     12.17
&   0.554   &      9.45 
\\

\midrule

\multicolumn{10}{l}{\textit{Multi-View Restoration}} \\
\midrule

VRT~\cite{liang2024vrt}
&    0.882   &     7.87  
&    3.226   &     10.68 
&    8.770  
&    0.750   &     6.98
&    0.558   &     8.96  
\\

FMA-Net~\cite{Youk_2024_CVPR}
&   1.319    &   4.45
&   3.623    &   11.92
&   8.653    
&   1.171    &   4.53
&   0.813    &   5.80
\\

VAE$_\text{MVD}$
&   \textbf{0.389}    &     \underline{11.63}
&   5.518    &     4.63
&   \underline{7.571}
&   \underline{0.314}    &    \underline{20.15}
&   0.636    &    6.21
\\


\textbf{GARD$_\text{BASE}$ (Ours)}
&  \underline{0.512}    &    \textbf{12.69}
&  \textbf{2.047}   &     \textbf{20.38}  
&  \textbf{6.304}
&  \textbf{0.232}     &     \textbf{23.41}
&  \textbf{0.347}     &     \textbf{21.60}  
\\

\bottomrule
\end{tabular}
}

\end{table*}
\begin{table*}[h]
\caption{\textbf{Quantitative depth estimation results.} We report AbsRel$\downarrow$ and $\delta_1\uparrow$ on the DA3 benchmark~\cite{depthanything3} using the DA3-BASE~\cite{depthanything3} checkpoint as the feed-forward 3D reconstruction model. The best result is highlighted in \textbf{bold} and the second best is \underline{underlined}.}
\label{supple_tab:da3_base_depth}
\centering

\resizebox{\linewidth}{!}{
\begin{tabular}{lccccc ccccc ccccc ccccc}
\toprule
\multirow{2}{*}{\textbf{Model}}
& \multicolumn{2}{c}{\textbf{HiRoom}}
& \multicolumn{2}{c}{\textbf{ETH3D}}
& \multicolumn{2}{c}{\textbf{DTU}}
& \multicolumn{2}{c}{\textbf{7Scenes}}
& \multicolumn{2}{c}{\textbf{ScanNet$++$}} \\
\cmidrule(lr){2-3} \cmidrule(lr){4-5} \cmidrule(lr){6-7} \cmidrule(lr){8-9} \cmidrule(lr){10-11}

& AbsRel$\downarrow$ & $\delta_1\uparrow$
& AbsRel$\downarrow$ & $\delta_1\uparrow$
& AbsRel$\downarrow$ & $\delta_1\uparrow$
& AbsRel$\downarrow$ & $\delta_1\uparrow$
& AbsRel$\downarrow$ & $\delta_1\uparrow$ \\
\midrule

\textit{Input Conditions} \\
\midrule

HQ Input
&    0.039   &       98.28
&    0.035   &      98.89
&    0.051   &      97.13
&    0.068  &       93.50
&    0.044   &       97.08
\\


LQ Input
&   0.195    &       77.75
&   0.124    &      84.75
&   0.057   &       95.66
&   0.144    &      79.19
&   0.123    &      85.07
\\

\midrule

\textit{Single-View Restoration} \\
\midrule


Restormer~\cite{zamir2022restormer}
&   0.202    &      76.56
&   0.147    &      80.91
&   0.057    &       95.71
&   0.119    &      84.47
&   \underline{0.112}    &      \underline{86.36}
\\


HI-Diff~\cite{chen2023hierarchical}
&   0.237    &       71.45
&   0.122    &       84.88
&   0.063    &       94.36
&   0.143    &       80.08
&   0.131    &       84.02
\\


InstructIR~\cite{conde2024high}
&   0.215    &       74.66
&   0.142    &       81.43
&   \underline{0.054}   &      \underline{95.94}
&   0.123    &       83.69
&   0.143    &       82.27
\\

MoCE-IR~\cite{zamfir2024complexityexperts}
&  0.186     &       79.12
&  0.124     &       85.07
&  \textbf{0.053}     &       \textbf{96.16}
&  0.132     &       81.52
&  0.122    &        85.51
\\

\midrule
\textit{Multi-View Restoration} \\
\midrule

VRT~\cite{liang2024vrt}
&   0.195    &      77.72
&   \underline{0.122}    &      \underline{85.24}
&   0.056    &      95.93
&   0.147    &      78.14
&   0.125    &      84.70
\\

FMA-Net~\cite{Youk_2024_CVPR}
&   0.251    &       69.54
&   0.137    &       82.12
&   0.063    &       95.13
&   0.178    &       73.26
&   0.159    &       79.07
\\

VAE$_\text{MVD}$
&    \underline{0.099}   &       \underline{92.80}
&    0.248   &       63.74
&    \underline{0.054}   &       95.85
&    \underline{0.106}   &       \underline{87.43}
&    0.160   &        77.57
\\


\textbf{GARD$_\text{BASE}$ (Ours)}
&   \textbf{0.079}    &   \textbf{95.04}
&   \textbf{0.063}    &   \textbf{95.54}
&   0.060    &   95.23
&   \textbf{0.081}   &    \textbf{91.60}
&   \textbf{0.059}    &   \textbf{95.76}
\\

\bottomrule
\end{tabular}
}
\end{table*}

\section{Implementation Details}

\subsection{Evaluation Metrics}
We follow the DA3~\cite{depthanything3} evaluation protocol for pose estimation and 3D reconstruction. For pose estimation, we report the area under the curve (AUC) of Relative Rotation Accuracy (RRA) and Relative Translation Accuracy (RTA) at thresholds of 5 and 30, reflecting strict and relaxed evaluation criteria. For 3D reconstruction, we report accuracy and completeness, and compute the Chamfer Distance (CD) as their average. We additionally report the F1-score under a distance threshold to assess overall reconstruction quality. For image restoration, we use PSNR and LPIPS to evaluate pixel-level fidelity and perceptual quality, respectively. PSNR measures reconstruction accuracy against the ground-truth image, while LPIPS assesses perceptual similarity in a deep feature space. Together, they evaluate both low-level fidelity and perceptual quality of the restored images. Higher PSNR and lower LPIPS indicate better restoration performance.

\subsection{RGB Image Decoder}
Our GARD framework jointly restores 3D scene geometry and high-quality RGB images by propagating the restored representations to their corresponding decoders. The underlying feed-forward 3D reconstruction model already incorporates a 3D geometry decoder $\mathcal{D}$ for scene geometry prediction. To reconstruct restored RGB images, we adopt the ViT-based RGB decoder $\mathcal{D}_{\text{rgb}}$ introduced in GLD~\cite{jang2026gld}. As the original RGB decoder was designed exclusively for the DA3-BASE backbone, we further adapt and fine-tune it for the DA3-GIANT backbone, which serves as the primary feed-forward 3D reconstruction model in our framework. Concretely, we initialize the RGB decoder using the released DA3-BASE checkpoint and introduce an additional DA3 linear projection adapter to bridge the feature dimensional discrepancy between the DA3-BASE and DA3-GIANT representations. The restored multi-scale features extracted from the four selected feature levels $\mathcal{M}=\{20, 28, 34, 40\}$ are concatenated and subsequently fed into the RGB decoder to reconstruct the restored RGB images. We refer readers to the original GLD work~\cite{jang2026gld} for additional architectural and implementation details.

\subsection{Training Details}

\paragraph{Training dataset.}
We train our model on two photo-realistic synthetic datasets: Hypersim~\cite{roberts2021hypersim} and TartanAir~\cite{wang2020tartanair}, which together cover diverse 3D environments. Hypersim provides high-fidelity indoor scenes with realistic lighting and materials, making it well-suited for learning appearance-aware geometric representations. In contrast, TartanAir includes both indoor and outdoor environments with dynamic camera trajectories, emphasizing temporal consistency and large viewpoint changes for multi-view geometry learning. By combining them, we leverage complementary strengths in visual realism and motion diversity, enabling more robust and generalizable geometric representations. Degraded images are generated by applying motion blur kernels using the codebase\footnote{https://github.com/LeviBorodenko/motionblur}. During training, we resize the input frames so that the longest side is 504 pixels and randomly select up to 4 views per iteration to train the model in various multi-view settings. 

\paragraph{Training hyperparameters.}
We train the GARD denoiser for 10 epochs with a global batch size of 8 on NVIDIA H100 GPUs, using the AdamW optimizer with a learning rate of $2 \times 10^{-4}$, $(\beta_1, \beta_2) = (0.9, 0.95)$, and no weight decay. All parameters are updated in full precision (fp32). We apply exponential moving average (EMA) with a decay rate of 0.9995 to stabilize training. The flow-matching objective uses a linear path with velocity prediction and logit-normal time sampling. Gradients are clipped to a maximum norm of 1.0.

\begin{figure}[h]
  \centering
  \includegraphics[width=\linewidth]{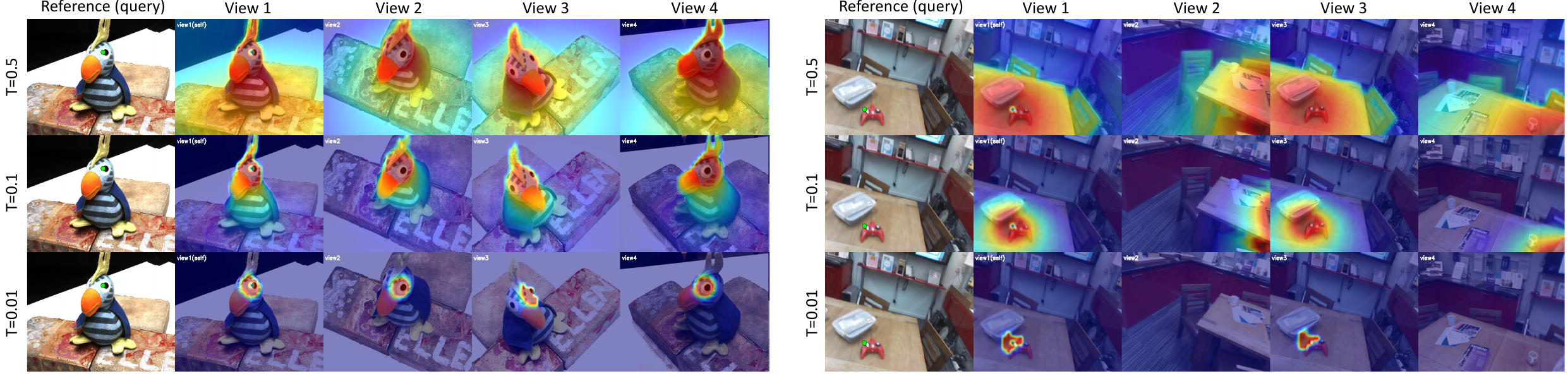}
\vspace{-15pt}
\caption{\textbf{Visualization of target correspondence maps} We visualize the effect of attention alignment training which augments the learning of the attention of the GARD denoiser.}
\label{fig:suppl_attn_target}
\end{figure}

\paragraph{Attention alignment.}
Fig.~\ref{fig:suppl_attn_target} visualizes the target correspondence map $\mathbf{A}^*$ under different temperature values $T$. To construct $\mathbf{A}^*$, we first estimate depth and camera pose from the clean high-quality (HQ) input views and unproject each pixel into a shared 3D world coordinate system. The resulting point cloud is spatially pooled to the ViT patch resolution (patch size 14) to align with the spatial token grid of the GARD denoiser $\mathcal{S}_{\theta}(\cdot)$. We then compute pairwise negative $\ell_2$ distances across all $VN$ patch tokens and apply a temperature-scaled softmax to obtain $\mathbf{A}^*$. Lower temperatures produce sharper and more spatially localized correspondence distributions, providing a stronger supervisory signal for the cross-entropy-based attention alignment loss $\mathcal{L}_{\text{attn}}$, whereas higher temperatures yield softer distributions across neighboring patches. For training the GARD denoiser, we used $T=0.01$ to construct the target correspondence map.

\begin{figure}[h]
  \centering
  \includegraphics[width=\linewidth]{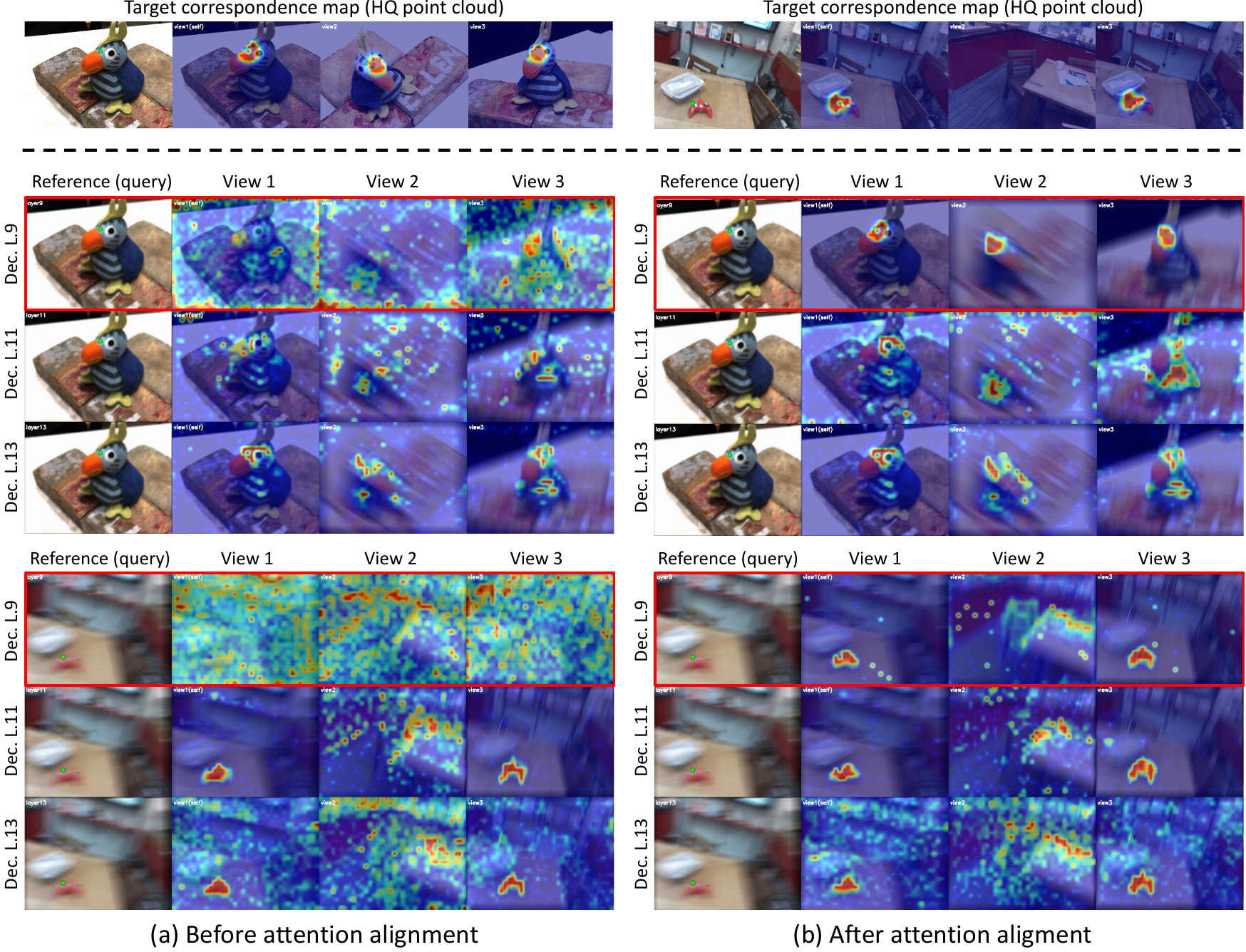}
\vspace{-10pt}
\caption{\textbf{Visualization of attention alignment} We visualize the effect of attention alignment training which augments the learning of the attention of the GARD denoiser.}
\label{fig:suppl_attn}
\end{figure}

Fig.~\ref{fig:suppl_attn} presents visualizations of the GARD denoiser attention maps before and after applying the attention alignment loss. Specifically, we first train the GARD denoiser using only the flow-matching objective and visualize the global attention maps from the wide decoder layers within the denoiser. While the 11th and 13th decoder layers exhibit relatively accurate attention toward the green query point, the 9th decoder layer fails to establish reliable geometric correspondences. Motivated by both this observation and the significance of wide decoder layers demonstrated in RAE~\cite{zheng2025diffusion}, we explicitly regularize the global attention weights of the $J$-th decoder layer ($J=9$) to encourage attention toward geometrically corresponding regions rather than spurious artifacts. As reported in Tab.~\ref{tab:abl_gard_training}, although attention alignment does not consistently improve performance under standard flow matching, combining it with interpolated flow matching yields additional gains. We attribute this improvement to the initialization from an LQ structural prior rather than pure noise, which enables more effective learning of geometrically consistent correspondences.

\paragraph{View selection strategy.}
To construct multi-view training samples, we adopt a view selection strategy based on the ground-truth camera poses provided in each dataset. Given a target number of views $V$, we retrieve neighboring frames using an expansion ratio that defines a larger candidate window to select the required number of views. For Hypersim, we use a \textit{near-camera} strategy since adjacent views often exhibit relatively large viewpoint changes. For every anchor frame, we first precompute the ranking between frame pairs based on their geometric overlap, and then select additional candidates to form the multi-view input. We use an expansion ratio of 2, which encourages stronger geometric consistency across selected views. In contrast, TartanAir consists of long video sequences with strong temporal continuity, where consecutive frames are often highly redundant. To increase viewpoint diversity during training, we employ a \textit{near-random} strategy with an expansion ratio of 4. Using this strategy, we first collect a larger set of nearby frames around the anchor frame and then randomly sample the remaining views from this candidate window. For training, we set the maximum number of training input views to $V=4$.

\subsection{Inference Details}
During inference, degraded multi-view images are first processed using the DA3-GIANT-1.1 multi-view encoder~\cite{depthanything3}. We extract the intermediate geometry-aware representation from the $K=18$-th encoder layer and use it as input to the proposed GARD denoiser. The denoiser restores the degraded representation by modeling the flow field in the latent feature space. To perform denoising, we solve the corresponding probability flow ordinary differential equation (ODE) using a first-order Euler solver with 50 discretized sampling steps. The restored representation is subsequently propagated through the remaining layers of the multi-view encoder to obtain refined multi-scale features. From these features, four feature levels $\mathcal{Z}_{\text{res}} = {\mathbf{z}_{\text{res}}^{l}}_{l \in \mathcal{M}}$ are selected, where $\mathcal{M}=\{20, 28, 34, 40\}$ for DA3-GIANT-1.1 and $\mathcal{M}=\{6, 8, 10, 12\}$ for DA3-BASE. These features are provided to the geometry decoder and RGB decoder~\cite{jang2026gld} for downstream 3D reconstruction and image restoration, respectively. Overall, the proposed GARD framework jointly reconstructs high-quality multi-view images and accurate 3D geometry in a single forward pass, eliminating the need for separate restoration and reconstruction stages.

\section{Additional Qualitative Results}
\begin{figure}[h]
    \centering
   \includegraphics[width=\textwidth]{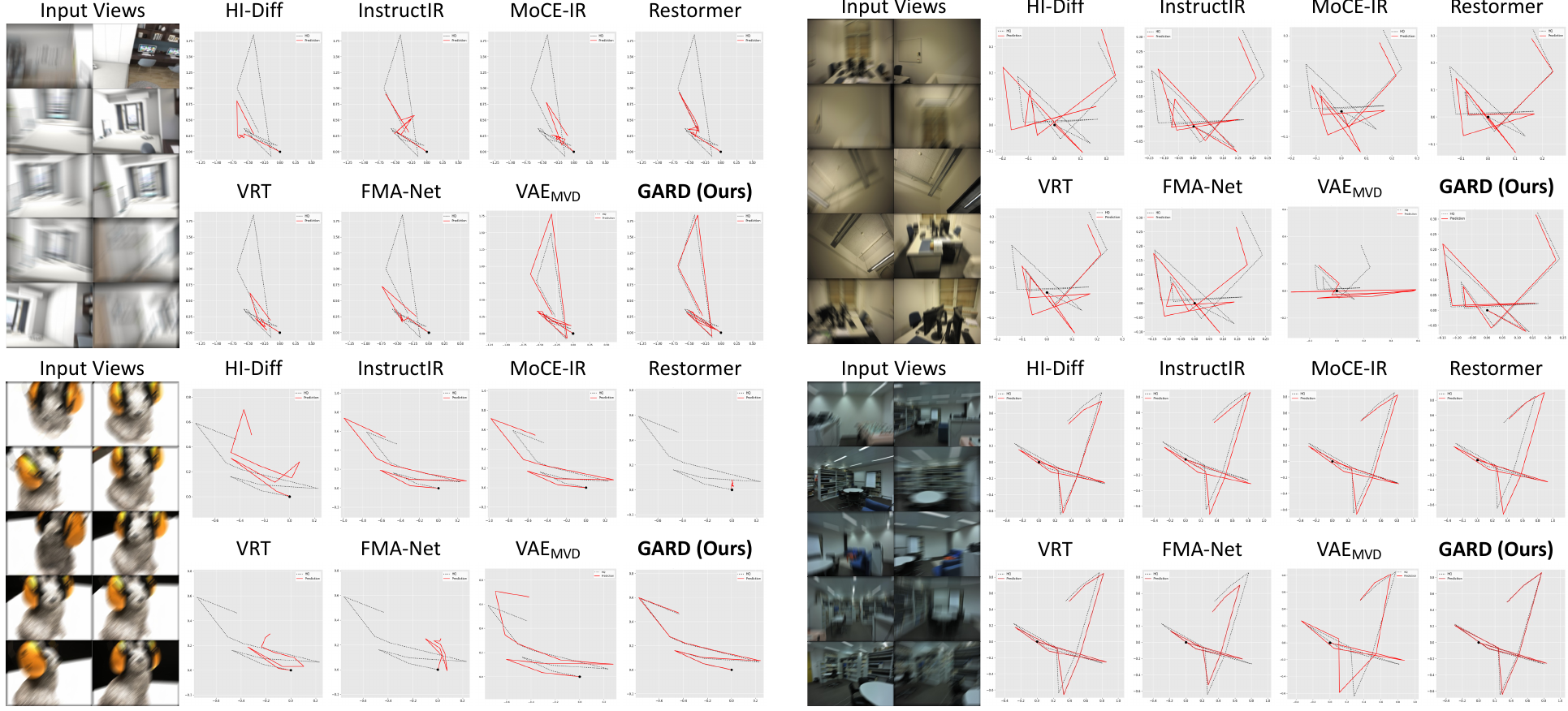}
    \caption{
    \textbf{Qualitative camera pose estimation on the DA3 benchmark~\cite{depthanything3}.} We visualize the top-down camera trajectory results for ten input views. The black dot indicates the starting camera point. Please zoom in for clearer visualization.
    }
    \label{fig:fig_suppl_pose}
\end{figure}

\begin{figure}[h]
    \centering
   \includegraphics[width=\linewidth]{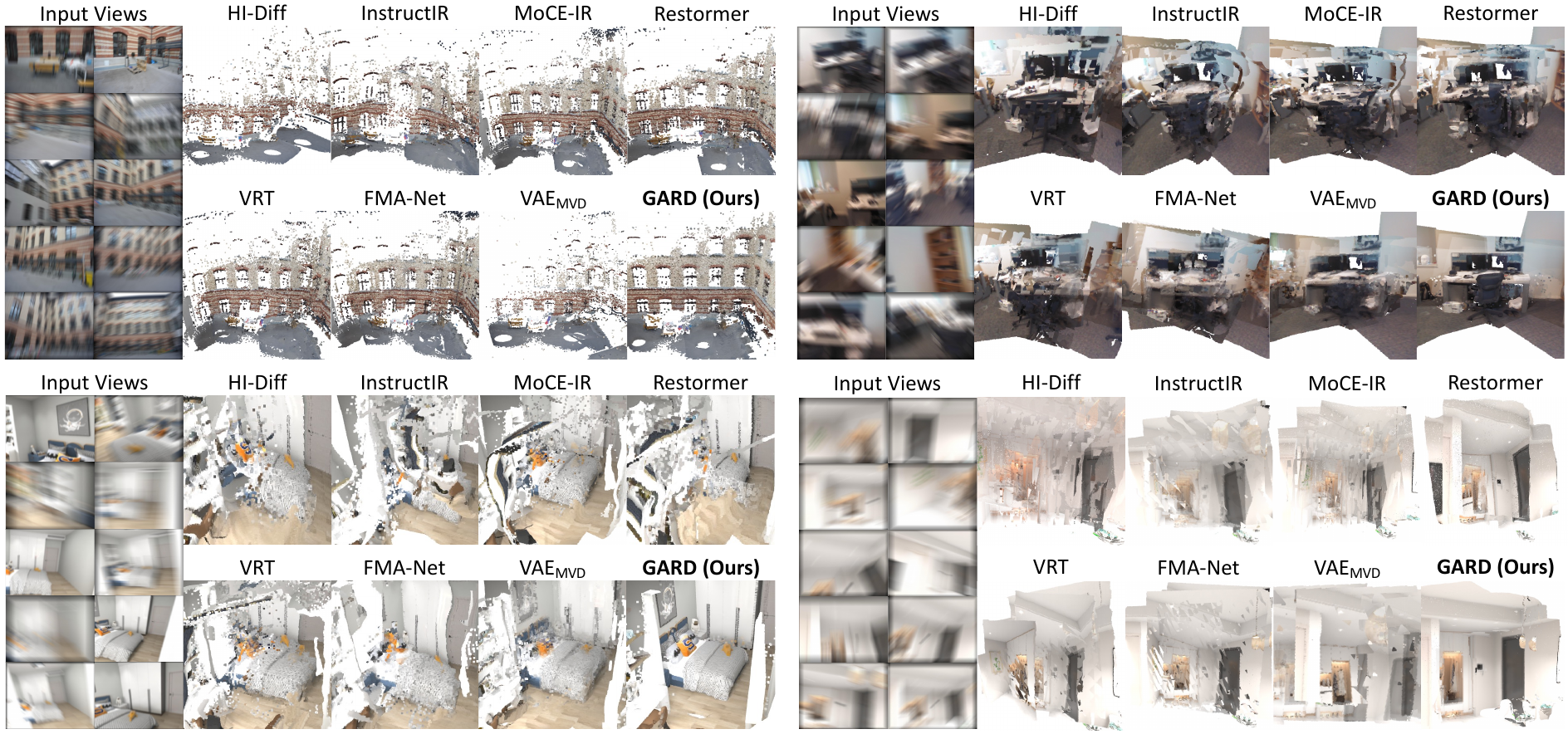}
    \caption{
    \textbf{Qualitative 3D reconstruction results on the DA3 benchmark~\cite{depthanything3}.} We visualize the 3D reconstruction point cloud results for ten input views. Please zoom in for clearer visualization.}
    \label{fig:fig_supple_recon}
\end{figure}

\begin{figure}[h]
    \centering
   \includegraphics[width=\textwidth]{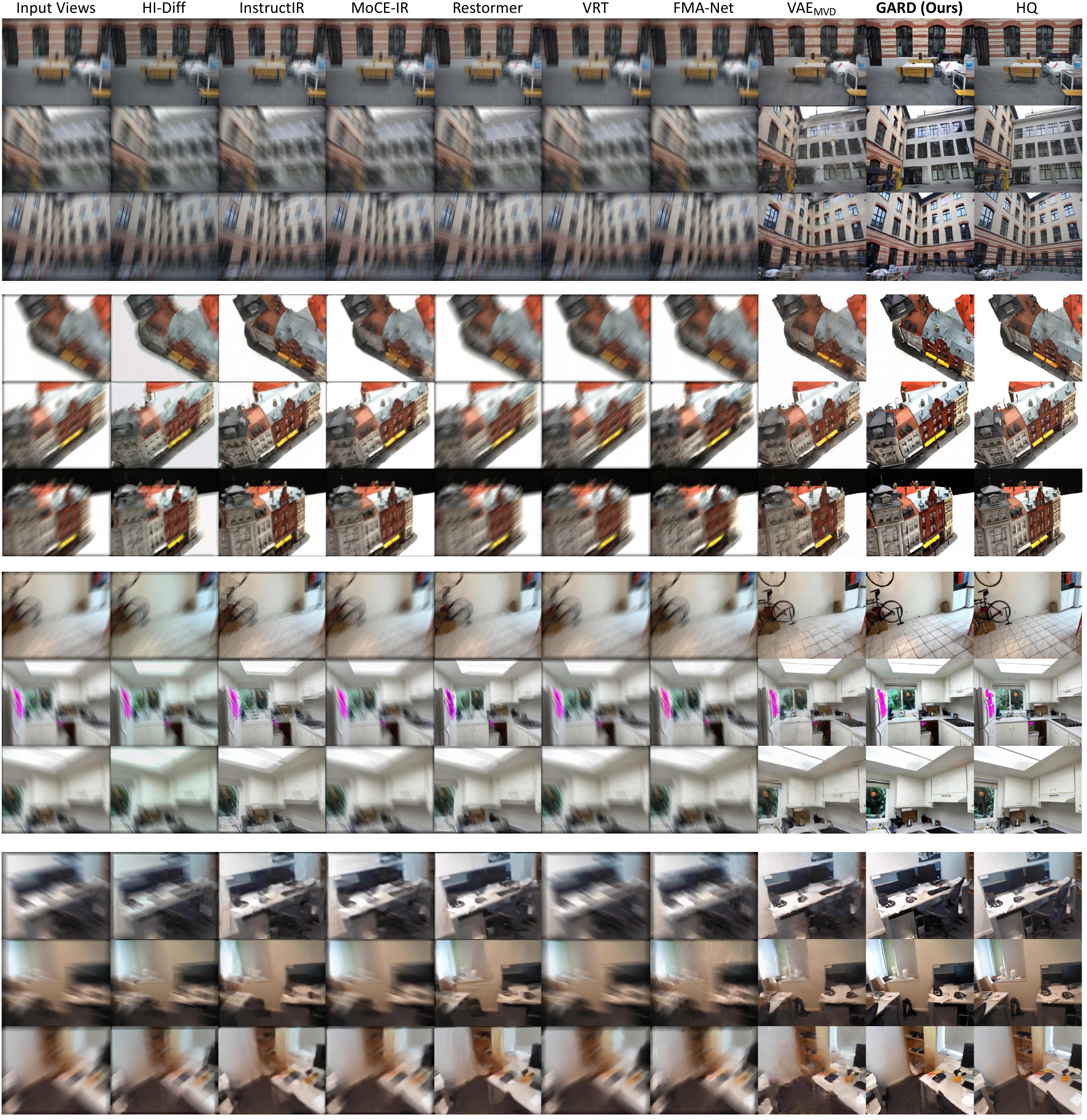}
    \caption{
    \textbf{Qualitative image restoration results on the DA3 benchmark~\cite{depthanything3}.} We visualize three selected views out of ten input views for each dataset. Please zoom in for clearer visualization.
    }
    \label{fig:fig_suppl_rgb_recon}
\end{figure}

\begin{figure}[h]
    \centering
    \includegraphics[width=\linewidth]{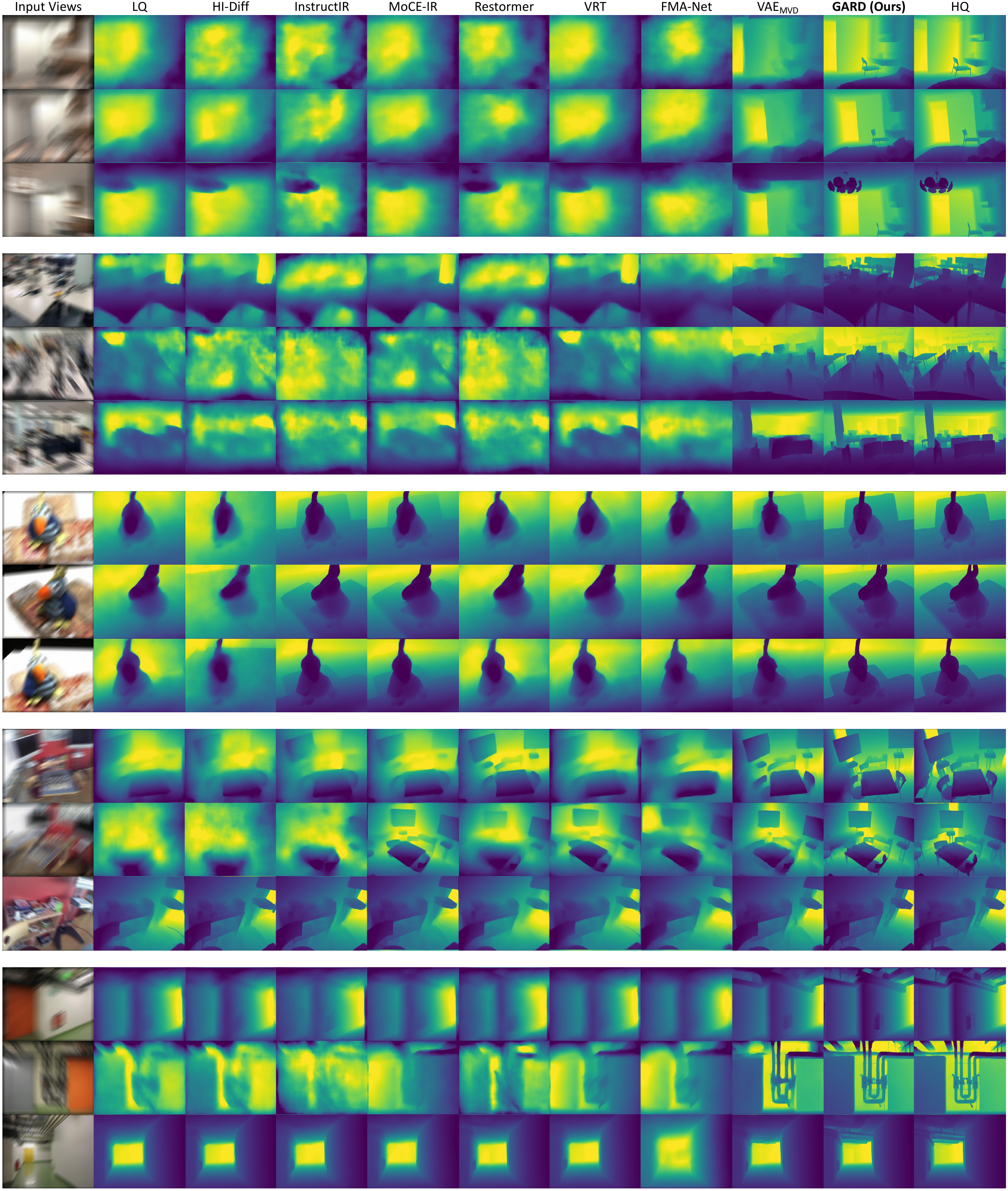}
    \caption{
    \textbf{Qualitative depth estimation results on the DA3 benchmark~\cite{depthanything3}.} We visualize three selected views out of ten input views for each dataset. Please zoom in for clearer visualization.
    }
    \label{fig:fig_suppl_depth}
\end{figure}

We present additional qualitative evaluation results for pose estimation, 3D reconstruction, image restoration and depth estimation in Fig.~\ref{fig:fig_suppl_pose}, Fig.~\ref{fig:fig_supple_recon}, Fig.~\ref{fig:fig_suppl_rgb_recon}, and Fig.~\ref{fig:fig_suppl_depth} respectively.

\clearpage
\vspace*{\fill}

\vfill

\vspace*{\fill}
\clearpage

\clearpage
\vspace*{\fill}

\vspace*{\fill}
\clearpage

\clearpage
\vspace*{\fill}
\vspace*{\fill}
\clearpage





\end{document}